%% file: emnlp2020-templates/main.tex
\pdfoutput=1

\documentclass[11pt]{article}

\usepackage{acl}
\usepackage{xspace} 
\usepackage{times}
\usepackage{latexsym}
\usepackage{enumitem}
\usepackage{tcolorbox}
\usepackage[T1]{fontenc}

\usepackage[utf8]{inputenc}
\usepackage{longtable}
\usepackage{microtype}

\usepackage{hyperref}
\usepackage{booktabs}
\usepackage{graphicx}
\graphicspath{{./imgs/}}
\usepackage{footmisc}
\usepackage{microtype}
\input{emnlp2020-templates/math-com}
\usepackage{arydshln}
\usepackage{subcaption}
\usepackage{caption}
\usepackage{array, makecell} %
\usepackage{footmisc}
\usepackage{alltt}
\usepackage{floatrow}
\usepackage{pifont}
\usepackage{tabularx}
\usepackage{adjustbox}
 \usepackage{multirow, caption}
 \usepackage{makecell}
\usepackage[table, dvipsnames]{xcolor} 
\usepackage{multirow, colortbl, caption}
\definecolor{lightblue}{RGB}{232, 244, 248}
\definecolor{lightpink}{RGB}{254, 238, 237}

\definecolor{bluelink}{RGB}{0,113,188}
\definecolor{greenlink}{RGB}{0,188,113}

\usepackage{tikz}

\usepackage{collcell}

\usepackage{etoolbox}

\newtoggle{inTableHeader}
\toggletrue{inTableHeader}
\newcommand*{\StartTableHeader}{\global\toggletrue{inTableHeader}}%

\newcommand{\dsworld}{DSAgentBench}

\let\OldTabular\tabular%
\let\OldEndTabular\endtabular%
\renewenvironment{tabular}{\StartTableHeader\OldTabular}{\OldEndTabular\StartTableHeader}%

\newcommand*{\MinNumber}{-1.0}%
\newcommand*{\MidNumber}{0.0} %
\newcommand*{\MaxNumber}{1.0}%

\newcommand{\ApplyGradient}[1]{%
  \iftoggle{inTableHeader}{#1}{
    \ifdim #1 pt > \MidNumber pt
        \pgfmathsetmacro{\PercentColor}{max(min(100.0*(#1 - \MidNumber)/(\MaxNumber-\MidNumber),100.0),0.00)} %
        \hspace{-0.33em}\colorbox{yellow!\PercentColor!blue}{#1}
    \else
        \pgfmathsetmacro{\PercentColor}{max(min(100.0*(\MidNumber - #1)/(\MidNumber-\MinNumber),100.0),0.00)} %
        \hspace{-0.33em}\colorbox{blue!\PercentColor!blue}{#1}
    \fi
  }}
\newcolumntype{R}{>{\collectcell\ApplyGradient}c<{\endcollectcell}}

\usepackage[nameinlink]{cleveref}
\crefformat{section}{\S#2#1#3} 
\crefname{algorithm}{Alg.}{Algs.}
\crefname{table}{Table}{Tables}
\crefformat{subsection}{\S#2#1#3}
\Crefname{equation}{Eq.}{Eqs.}
\Crefname{figure}{Figure}{Figures}

\usepackage[colorinlistoftodos,prependcaption,textsize=tiny]{todonotes}

\usepackage{soul}

\usepackage{float}

\usepackage{multirow}
\usepackage{hhline}

\usepackage{amssymb}   
\usepackage{stmaryrd}  

\definecolor{headerLavender}{RGB}{230, 230, 250} 
\definecolor{rowLightGray}{RGB}{245, 245, 245} 
\definecolor{rowCream}{RGB}{255, 250, 240} 
\definecolor{errorRed}{RGB}{255, 77, 77} 

\definecolor{chartqapro1}{RGB}{30,160,220} 
\definecolor{chartqapro2}{RGB}{50,200,100} 

\definecolor{dashqa1}{RGB}{63,81,181}   
\definecolor{dashqa2}{RGB}{0,188,212}   

\newcommand{\dashInteractqa}[1]{\textsc{
\textcolor{dashqa1}{Dashboard}\textcolor{dashqa2}{QA}}}

\definecolor{dsworld1}{RGB}{0,85,160}
\definecolor{dsworld2}{RGB}{220,60,50}
\newcommand{\DSAgentBench}[1][]{\textsc{\textcolor{dsworld1}{DS}\textcolor{dsworld2}{Agent}Bench}}

 \title{\protect\ DSAgentBench: Can Agents Automate End-to-End Data-Science Workflows in Real Computer Environments?} 

\author{
\textbf{Mizanur Rahman}\textsuperscript{\textdaggerdbl}
\thanks{Corresponding authors: \{mizanurr,enamulh\}@yorku.ca},
\textbf{Mohammed Saidul Islam}\textsuperscript{\textdaggerdbl}, \\
\textbf{Ridwan Mahbub}\textsuperscript{\textdaggerdbl},
\textbf{Md Tahmid Rahman Laskar}\textsuperscript{\textdaggerdbl}, \\
\textbf{Shafiq Joty}\textsuperscript{\textdollar,\textparagraph},
\textbf{Enamul Hoque Prince}\textsuperscript{\textdaggerdbl}\footnotemark[1]
\\[2pt]
\textsuperscript{\textdaggerdbl}York University \\
\textsuperscript{\textdollar}Nanyang Technological University \quad
\textsuperscript{\textparagraph}Salesforce AI Research
}

\begin{document}

\maketitle

\begin{abstract} 
Real-world data science involves long-horizon workflows that span data wrangling, exploration, modeling, visualization, and validation, and require coordinated use of tools such as notebooks, IDEs, terminals, browsers, and databases within real operating environments. Yet existing benchmarks lack real-computer interaction and do not evaluate whether agents can execute complete end-to-end data-science workflows in realistic computing environments, failing to capture the multi-stage, multi-tool nature of data-science practice. We introduce \DSAgentBench{}, the first benchmark to evaluate whether agents can automate full data-science workflows inside real computer environments. \DSAgentBench{} contains 275 diverse tasks covering the entire data-science life-cycle,  reflecting the complexity and tool coordination required in practice.  Each task requires grounding decisions in intermediate outputs and coordinated tool use, and includes a deterministic evaluator that verifies analytical correctness, visual outputs, and model performance rather than code-only execution. Our extensive experiments with 15 closed- and open-source models show that even the strongest agent, Claude-4.6-Sonnet, achieves only 56.70\% task success, while all open-source agents remain below 1\%, frequently failing at tool orchestration, OS grounding, and multi-step reasoning. These results reveal a substantial capability gap between current agentic systems and real data-science workflows, positioning \DSAgentBench{} as a foundation for developing grounded, verifiable, autonomous data-science agents. We release \DSAgentBench{} at \url{https://github.com/vis-nlp/DSAgentBench}.



\end{abstract}

\section{Introduction}
\begin{figure}[t!]
    \includegraphics[width=\textwidth]{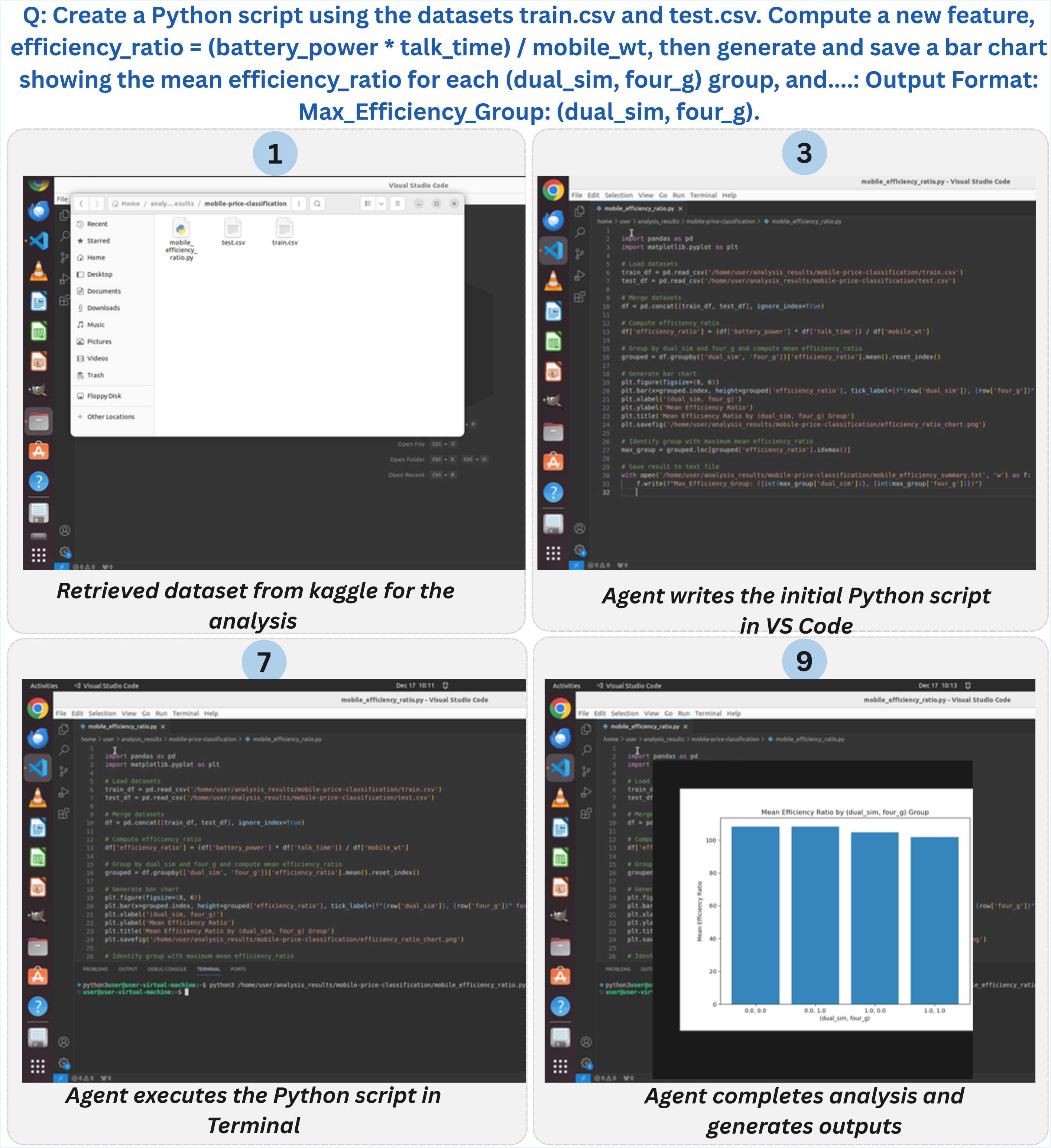}
    \caption{Example agent workflow task in \DSAgentBench{}, showing selected stages from a multi-step workflow: the agent retrieves data, executes code, and produces the final outputs. The illustrated trajectory is generated by GPT-4o.    
    }
    \label{fig:dsworld-overview-example}
\end{figure}


Data science drives critical decisions across industries by transforming raw data into actionable insights, from fraud detection to drug discovery and climate modeling \cite{sarker2021data,adeniran2024role, donoho201750}. Achieving such outcomes requires a broad skill set that includes programming, statistical reasoning, visualization literacy, domain knowledge, machine learning, and effective communication \cite{cao2017data}.  
  Crucially, real-world data science practice extends far beyond generating correct code: analysts must inspect heterogeneous datasets, scrape 
  information from web sources, manage dependencies,  execute and debug scripts, generate visualizations, and iteratively refine predictive models \cite{dale2022data,wickham2023r}. These workflows span multiple tools~\cite{zhang2020data, hong2025data}, including database s, notebooks, IDEs, terminals, and web interfaces, and demand continuous analytical reasoning within computing environments (Figure \ref{fig:dsworld-overview-example}). 

Recent advances in large language models have shown promise in automating portions of the data science workflow \cite{rahman2025llm,jiang2024survey}. Models can convert natural language instructions into executable Python code, infer dataset schemas, generate statistical summaries, and propose analysis strategies \cite{liu2023your,lei2024spider,chen2024viseval}. However, these abilities address only a limited subset of real-world data science practice. 
Producing syntactically and functionally 
correct code is fundamentally different from operating autonomously within real computing environments, which requires navigating file systems, coordinating tools, interpreting errors, and refining analyses based on intermediate outputs \cite{huang2024code}. Industry reports reinforce this gap, noting that current AI systems struggle to reliably execute multi-step, cross-tool analytical workflows despite strong performance on isolated tasks \cite{openai2025computer}. This gap raises a central question: \textit{
Can AI agents perform long-horizon reasoning to autonomously execute end-to-end data-science workflows in real computer environments?}




Vision–language–model agents offer a promising path toward such autonomy by enabling interaction with computer environments through visual observations and GUI-based actions \cite{wang2024gui,tang2025survey}. Benchmarks such as OSWorld \cite{xie2024osworld}, WebArena \cite{zhou2023webarena}, VisualWebArena \cite{koh2024visualwebarena}, ScreenSpot-Pro \cite{li2025screenspot}, and  DashboardQA \cite{kartha2025dashboardqa} show that agents can control applications and execute multi-step tasks across web and mobile environments. However, these benchmarks primarily evaluate general computer interaction, not whether agents can reason through and execute end-to-end data-science workflows.



Conversely, existing data-science benchmarks evaluate analytical capabilities without real system interaction. Most benchmarks, including DS-1000 \cite{lai2023ds}, DABStep \cite{egg2025dabstep}, MLAgentBench \cite{huang2023mlagentbench}, DSBench \cite{jing2024dsbench}, and DSEval \cite{zhang2024benchmarking}  evaluate code correctness through static execution, without 
requiring agents to launch applications, navigate file systems, manage dependencies, or interact with terminals and IDEs. DA-CODE \cite{huang2024code} moves closer to realistic workflows by evaluating task planning and multi-file analysis, but remains confined to a sandboxed notebook environment without operating-system access or cross-tool coordination.  Consequently, existing benchmarks test whether code runs in isolation, not whether agents can autonomously operate systems and refine analyses through real tool interaction. Moreover, existing benchmarks evaluate isolated skills, not end-to-end data-science workflows from data acquisition to validation (Table~\ref{tab:dsworld_positioning}).

To address this gap, we propose  \DSAgentBench{}, a data-science environment 
that enables agents to perform real analytical workflows inside a 
functioning operating system. We leverage the OSWorld 
framework \cite{xie2024osworld} and augment it with tools essential for data science, including Jupyter 
Notebook, automated access to external data sources (Kaggle API, OpenML), 
and SQLite databases. 
The environment supports task-specific configuration that automatically prepares datasets and enables agents to carry out end-to-end data-science workflows across multiple tools, from data acquisition to analysis and visualization.
 Using this environment, we construct 275 long-horizon, human-authored tasks comparable in scale to existing OSWorld-level benchmarks \cite{xie2024osworld}, each paired with a deterministic evaluator that verifies analytical correctness rather than surface-level code execution. 
 Our evaluation shows that even the strongest agent, Claude-4.6-Sonnet, achieves only 56.70\% task success, while all open-source agents remain below 1\%, exposing major limitations in grounding, tool orchestration, and long-horizon reasoning. 
 

In summary, our contributions are:
\Ni \textbf{\dsworld{}}, the first benchmark for evaluating autonomous data-science workflows inside real operating systems, covering the full data-science lifecycle.
\Nii \textbf{An extension of OSWorld} that enables interaction with core data-science tools and external data sources (e.g., Kaggle, OpenML, SQLite).
\Niii \textbf{Deterministic, execution-based evaluation} that assesses analytical correctness, visualization outputs, and model performance beyond code-only execution, along with an extensive evaluation of 15 open- and closed-source agents.
\Niv \textbf{In-depth analysis and ablations} that identify key limitations in grounding, planning, and analytical reasoning and outline directions for future agent development.

\begin{table*}[t!]
\centering
\renewcommand{\arraystretch}{1.2}
\small
\setlength{\tabcolsep}{4pt}

\caption{\textbf{Comparison of \dsworld{} with existing benchmarks across unified agent and data-science capabilities.} \ding{51} = Yes, \ding{115} = Partial, \ding{55} = No.}

\label{tab:dsworld_positioning}

\resizebox{0.97\textwidth}{!}{
\begin{tabular}{l
c c c c c c c c c c}
\toprule
\textbf{Benchmark} &
\makecell{\textbf{Full OS} \\ \textbf{Interaction}} &
\makecell{\textbf{Web Tools} \\ \textbf{\& Browsers}} &
\makecell{\textbf{Terminal +} \\ \textbf{GUI Control}} &
\makecell{\textbf{Cross-App} \\ \textbf{Usage}} &
\makecell{\textbf{Intermediate} \\ \textbf{State}} &
\makecell{\textbf{Controlled} \\ \textbf{Exec Env}} &
\makecell{\textbf{Multimodal} \\ \textbf{Support}} &
\makecell{\textbf{Data-Science} \\ \textbf{Tasks}} &
\makecell{\textbf{Visualization} \\ \textbf{Evaluation}} &
\makecell{\textbf{Environment} \\ \textbf{Scalability}} \\
\midrule

\rowcolor[HTML]{F2F2F2} \textbf{HumanEval} \cite{chen2021evaluating} &
\ding{55} & \ding{55} & \ding{55} & \ding{55} &
\ding{55} & \ding{55} &
\ding{55} &
\ding{55} &
\ding{55} &
\ding{55} \\

\rowcolor[HTML]{F2F2F2} \textbf{KRAMABench} \cite{lai2025kramabench} &
\ding{55} & 
\ding{55} & 
\ding{55} & 
\ding{55} & 
\ding{51}  & 
\texttt{code} & 
\ding{55} & 
\ding{115} & 
\ding{55} & 
\ding{55} \\ 

\rowcolor[HTML]{F2F2F2} \textbf{DS-1000} \cite{lai2023ds} &
\ding{55} & 
\ding{55} & 
\ding{55} & 
\ding{55} & 
\ding{55} & 
\ding{55} & 
\ding{55} & 
\ding{115} & 
\ding{55} & 
\ding{55} \\ 

\rowcolor[HTML]{F2F2F2} \textbf{DABStep} \cite{egg2025dabstep} &
\ding{55} &
\ding{55} &
\ding{55} &
\ding{55} &
\ding{51} &
\texttt{code} &
\ding{55} &
\ding{115}  &
\ding{55} &
\ding{55}  \\

\rowcolor[HTML]{F2F2F2} \textbf{MLAgentBench} \cite{huang2023mlagentbench} &
\ding{55} & 
\ding{55} & 
\ding{55} & 
\ding{55} & 
\ding{55} & 
\ding{55} & 
\ding{55} & 
\ding{115} & 
\ding{55} & 
\ding{55} \\ 

\rowcolor[HTML]{F2F2F2} \textbf{DSBench} \cite{jing2024dsbench} &
\ding{55} & 
\ding{55} & 
\ding{55} & 
\ding{55} & 
\ding{55} & 
\ding{55} & 
\ding{51} & 
\ding{115} & 
\ding{55} & 
\ding{55}\\ 

\rowcolor[HTML]{F2F2F2} \textbf{DSEval} \cite{zhang2024benchmarking} &
\ding{55} & 
\ding{55} & 
\ding{55} & 
\ding{55} & 
\ding{55}  & 
\ding{51} & 
\ding{55} & 
\ding{115} & 
\ding{55} & 
\ding{55} \\ 

\rowcolor[HTML]{F2F2F2} \textbf{ARCADE} \cite{yin2023natural} &
\ding{55} & 
\ding{55} & 
\ding{55} & 
\ding{55} & 
\ding{55} & 
\ding{55}& 
\ding{55} & 
\ding{115} & 
\ding{55} & 
\ding{55}\\ 

\rowcolor[HTML]{F2F2F2} \textbf{DA-CODE} \cite{huang2024code} &
\ding{55} & \ding{55} & \ding{55} & \ding{55} &
\ding{55}  & 
\texttt{code} &
\ding{55} &
\ding{115} &
\ding{115} &
\ding{115} \\

\rowcolor[HTML]{F2F2F2} \textbf{OSWORLD} \cite{xie2024osworld} &
\ding{51} & \ding{51} & \ding{51} & \ding{51} &
\ding{51} & \ding{51} &
\ding{51} &
\ding{55} &
\ding{55} &
\ding{51} \\

\midrule
\rowcolor[HTML]{E5F1FB} \textbf{\dsworld{} (Ours)} &
\ding{51} & \ding{51} & \ding{51} & \ding{51} &
\ding{51} & \ding{51} &
\ding{51} &
\ding{51} &
\ding{51} &
\ding{51} \\

\bottomrule
\end{tabular}}
\vspace{-2mm}
\end{table*}

\section{Related Work}

\paragraph{Data Science Benchmarks and Agents.} Early program synthesis benchmarks such as HumanEval \cite{chen2021evaluating} evaluate functional code correctness using unit tests and contain no data analysis tasks. More recent data-science benchmarks capture richer analytical capabilities but remain limited to isolated stages of the workflow. 
For example, MLAgentBench \cite{huang2023mlagentbench}, DABStep \cite{egg2025dabstep}, DSBench  \cite{jing2024dsbench}, DS-1000 \cite{lai2023ds}, and DSEval \cite{zhang2024benchmarking} focus on subsets of data manipulation, modeling, or reasoning, while visualization benchmarks such as ChartQA \cite{masry2022chartqa}, Text2Vis \cite{rahman2025text2vis}, and VisEval \cite{chen2024viseval} evaluate chart generation and visual reasoning without involving data wrangling or iterative analysis. Benchmarks targeting more complex pipelines, including DA-CODE \cite{huang2024code}, KRAMABench \cite{lai2025kramabench}, and ARCADE \cite{yin2023natural}, still operate in static, sandboxed environments without operating-system interaction. Consequently, existing benchmarks test code generation or isolated reasoning rather than autonomous execution of end-to-end data-science workflows. 
\dsworld{} fills this gap by enabling agents to execute complete data-science workflows inside a functioning operating system, requiring coordinated reasoning, tool orchestration, and environment interaction (Tab. \ref{tab:dsworld_positioning}). 

Recent work has also explored large language models as agents for data science by generating  code~\cite{qiao2023taskweaver, Guo-2024-DS-agent, li2024autokaggle} or managing relevant context across tasks~\cite{hong2025data}. However, these approaches are typically evaluated through case studies or controlled settings, making it difficult to assess robustness, long-horizon reasoning, and tool coordination in realistic computer environments. 

\paragraph{GUI and Computer-Control Agents}
Vision-language agents have enabled autonomous computer control through multimodal observation and GUI interaction. Early work focused on web environments: WebShop \cite{yao2022webshop} and Mind2Web \cite{deng2023mind2web} evaluated navigation and instruction following on simulated and real websites, while WebArena \cite{zhou2023webarena} and VisualWebArena \cite{koh2024visualwebarena} introduced realistic web applications requiring visual grounding and multi-step planning. OSWorld \cite{xie2024osworld} extended this line of work to full desktop control across Ubuntu, Windows, and macOS, where agents operate applications using keyboard and mouse actions guided by screenshots and accessibility trees. 
Additional benchmarks target mobile interfaces (AndroidWorld \cite{rawles2024androidworld}), office productivity (OfficeBench \cite{wang2024officebench}), and UI grounding (ScreenSpot-Pro \cite{li2025screenspot}). Recent work further improves desktop grounding through instruction-tuned models (CogAgent \cite{hong2024cogagent}, ShowUI \cite{lin2025showui}, Ferret-UI \cite{you2024ferret}), grounding benchmarks and agents (OS-ATLAS \cite{wu2024atlas}, GroundCUA \cite{feizi2025grounding}, JEDI \cite{xie2025scaling}), and reinforcement-learning-based agents (GUI-R1 \cite{luo2025gui}, GUI-G2 \cite{tang2025gui}, InfiGUI-G1 \cite{liu2025infigui}). 

These benchmarks establish core infrastructure for OS-level autonomy but primarily evaluate general computing operations, such as application use and interface navigation, rather than analytical reasoning. They assess whether agents can operate computers rather than function as data scientists who load data, analyze results, train models, visualize findings, and debug workflows.

\section{ \DSAgentBench{} Benchmark}

\input{emnlp2020-templates/Dataset}
\section{Methodology}
\subsection{Environment Architecture}  
\label{env}
\begin{figure}[t!]
    \includegraphics[width=.98\textwidth]{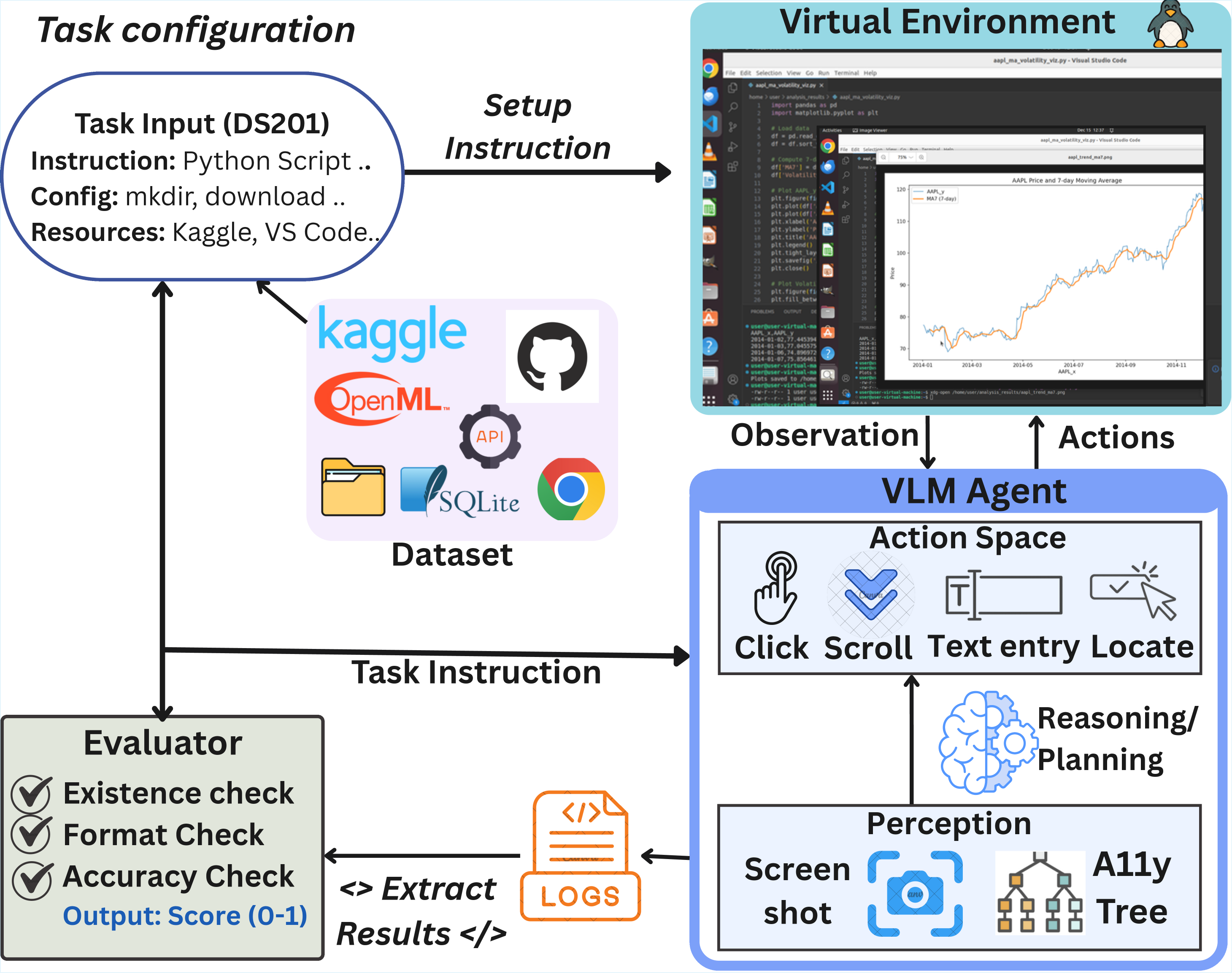}
\caption{
\DSAgentBench{} execution workflow: the agent perceives screenshots and the A11y tree, executes GUI-based actions, and is evaluated by a custom script.
}
    \label{fig:dsworld-overview}
\end{figure}

\vspace{-1mm}
\DSAgentBench{} extends OSWorld \cite{xie2024osworld} to provide a realistic execution environment for end-to-end analytical workflows (Figure \ref{fig:dsworld-overview}). The environment runs Ubuntu OS with Python and common data-science libraries pre-installed, and captures screenshots at a resolution of 1920$\times$1080 to provide clear visual context. Visual Studio Code and Jupyter Notebook are available for script development and interactive analysis, and Chrome is included for accessing documentation or external data when needed. The environment supports automated data retrieval through the Kaggle API, 
OpenML, direct URLs, and SQLite databases. 

Formally, the environment is represented as a tuple $(\mathcal{O}, \mathcal{A}, T)$, where $\mathcal{O}$ is the observation space, $\mathcal{A}$ is the action space, and $T$ is the transition function. At each timestep $t$, the agent receives an observation $o_t \in \mathcal{O}$, selects an action $a_t \in \mathcal{A}$, and obtains the next observation $o_{t+1}$ after the system transitions to state $s_{t+1} = T(s_t, a_t)$.

\textbf{Observation Space}: At each timestep $o_t$, the agent receives one of two modalities: \textit{(i)} a 1920$\times$1080 \textbf{Screenshot} of the desktop interface, providing pixel-level visual context; or \textit{(ii)} an integrated \textbf{Screenshot + Accessibility Tree (A11y)} modality that augments the screenshot with structured UI metadata, including element roles, accessible names, bounding boxes, and interaction states, extracted via AT-SPI
\footnote{\url{https://docs.gtk.org/atspi2/}}. 
These two settings allow us to study how structured UI metadata complements visual observations for grounding in data science workflows. Observations are captured after each action with sufficient delay to ensure that windows and outputs have fully stabilized.

\textbf{Action Space}: Agents interact with the environment through a structured action space $\mathcal{A}$ supporting the core operations required for data-science workflows. 
Actions include \textit{(i)} mouse-based GUI interactions (e.g., clicking, dragging, scrolling), \textit{(ii)} keyboard input for text entry and shortcut-based control (e.g., navigating IDE interfaces, code execution and file operations), and \textit{(iii)} meta actions including  \texttt{WAIT} to pause execution and allow rendering, \texttt{DONE} to signal task completion, and \texttt{FAIL} to 
terminate execution with an error. Each action executes atomically, after which the environment updates the desktop state and returns the next observation $o_{t+1}$, enabling realistic human–computer interaction across diverse analytical workflows.

All agents operate through a standard GUI action space to evaluate end-to-end desktop competence without relying on privileged task-specific APIs. This ensures fair comparison while still supporting terminal, CLI, and script-based workflows via typical interactions, reflecting how data science is performed in real computing environments.

\subsection{Task Execution Workflow}
Each task begins with environment initialization, loading a clean virtual-machine snapshot and applying task-specific setup steps (e.g., directory creation, dataset retrieval, application launch) to produce the initial observation. 
The agent then follows an iterative perception–action loop: at each timestep, it receives the current observation, selects an action based on the task instruction and interaction history, executes the action in the environment, and observes the updated desktop state. Execution ends when the agent signals completion via \texttt{DONE}, explicitly aborts with \texttt{FAIL}, or reaches the maximum step budget of 15 actions (see example Figure~\ref{fig:agent_interaction}). All observations, actions, and timestamps are logged throughout execution to enable reproducible evaluation, trajectory analysis, and failure diagnosis. Further details on implementation and evaluation prompts are provided in Appendix~\Cref{app:method,app:evaluation}, respectively.

\section{Evaluation}

\input{emnlp2020-templates/experiment}
\input{emnlp2020-templates/Evaluation}
\section{Conclusion}
We present \DSAgentBench{}, the first benchmark for assessing whether agents can automate end-to-end data-science workflows inside real operating systems.  Unlike prior benchmarks that focus on isolated code generation or generic GUI interaction, our benchmark requires agents to plan and execute long-horizon, multi-tool workflows spanning data acquisition, analysis, and modeling. Our experiments reveal substantial limitations in current agents, with even the strongest systems achieving low success rates on these tasks. Through systematic evaluation and analysis, we identify key challenges in grounding, tool orchestration, and iterative analytical reasoning. Failure patterns show that DSAgentBench jointly evaluates desktop grounding and data-science reasoning via real tool use, state management, and correct artifact generation.
We hope \DSAgentBench{} accelerates progress toward  more robust, and autonomous agents capable of performing real-world data science.

\section*{Limitations}

First, the open-source agents in our evaluation stack do not currently support A11y Tree observations, so we evaluate them under the screenshot-only setting. However, for closed-source agents, we report both screenshot-only and screenshot+A11y Tree results, where A11y provides only modest overall gains.


Second, our detailed error analysis is based on 604 manually inspected trajectories from closed-source models and 150 from open-source models. While this constitutes a substantial qualitative evaluation, it still represents a subset of the full benchmark, and rare failure patterns may not be fully captured.

Finally, our evaluation of visualization-heavy tasks focuses on final artifact quality. The evaluators verify required outputs, data mappings, labels, and task-specific correctness, overall visual clarity and semantic alignment. 



\section*{Ethical Considerations}

For the \DSAgentBench{} benchmark, we applied strict data governance and licensing constraints during dataset construction. All tabular and structured datasets were collected exclusively from openly licensed platforms, specifically GitHub, Kaggle, and OpenML. We included only datasets that are either MIT-licensed or released under clearly permissive open-source or public-use licenses and explicitly suitable for research, redistribution, and modification as specified by their providers. Datasets with ambiguous, restrictive, or missing license information were systematically excluded. This filtering ensures that all components of \DSAgentBench{}  comply with licensing requirements and can be safely used, shared, and extended by the research community. Finally, we used AI-based writing assistants only to improve the presentation of the paper.

\section*{Acknowledgements}
This research was supported by the Natural Sciences and Engineering Research Council (NSERC),
Canada, Canada Foundation for Innovation, Compute Canada, and the CIRC grant on Inclusive and
Accessible Data Visualizations and Analytics.



\bibliography{emnlp2020-templates/dashInteractQA}
\newpage
\input{emnlp2020-templates/Appendix}

\end{document}

%% file: emnlp2020-templates/math-com.tex
\usepackage{amsmath}
\usepackage{amsfonts,bm}
\usepackage{xspace}

\newcommand{\Ni}{({\em i})~}
\newcommand{\Nii}{({\em ii})~}
\newcommand{\Niii}{({\em iii})~}
\newcommand{\Niv}{({\em iv})~}

\definecolor{mypink3}{cmyk}{0, 0.7808, 0.4429, 0.1412}

\makeatletter   
\newcommand{\sveryshortarrow}[1][3pt]{\mathrel{%
    \vcenter{\hbox{\rule[-.5\fontdimen8\scriptfont3]
               {\scriptratio\dimexpr#1\relax}{\fontdimen8\scriptfont3}}}%
   \mkern-4mu\hbox{\let\f@size\sf@size\usefont{U}{lasy}{m}{n}\symbol{41}}}}
\makeatother

\def\eqref#1{equation~\ref{#1}}

\def\1{\bm{1}}

\def\m1{{\bm{1}}}

\DeclareMathAlphabet{\mathsfit}{\encodingdefault}{\sfdefault}{m}{sl}
\SetMathAlphabet{\mathsfit}{bold}{\encodingdefault}{\sfdefault}{bx}{n}



%% file: emnlp2020-templates/Dataset.tex
\begin{figure*}[t!]
    \centering
    \includegraphics[width=.97\textwidth]{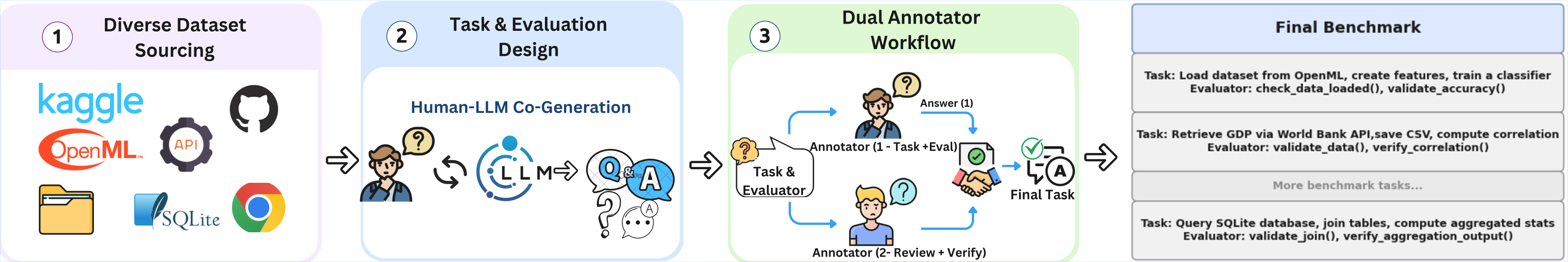} 
    \caption{Benchmark Construction Pipeline: We source heterogeneous real-world datasets (1); design tasks and evaluators through human–LLM collaboration with pre- and post-execution configurations (2); and apply dual-annotator review and verification (3). The result is a curated benchmark of reproducible tasks with complete configurations and deterministic evaluation scripts.}    
    \label{fig:dsworld_pipeline}
\end{figure*}

This section introduces \DSAgentBench{}, a diverse set of problems, and describes its data collection and annotation pipeline. 
\subsection{Problem Formulation}
\vspace{-1mm}
\DSAgentBench{} frames data science tasks as an autonomous, long-horizon data-driven decision-making problem in real computing environments. Formally, the benchmark consists of a set of tasks $\mathcal{T} = \{ (\mathcal{C}_i, \mathcal{I}_i, \mathcal{V}_i) \}_{i=1}^{N}$, where $\mathcal{C}_i$ is a task configuration defining the initial system setup, $\mathcal{I}_i$ is a natural language instruction describing the analytical objective, and $\mathcal{V}_i$ is a deterministic Python evaluator.

The task configuration $\mathcal{C}_i$ defines the initialized operating-system state, including available datasets, file system structure, installed libraries, and applications (e.g., IDEs, 
terminals, browsers), as well as optional initialization or cleanup procedures. An agent interacts with the environment through multimodal observations and actions to execute an end-to-end data-science workflow. DSAgentBench evaluates final outcomes rather than prescribing a fixed workflow path: agents may use any available applications as long as outputs satisfy the deterministic evaluator.
A task is 
successful if and only if the agent’s outputs satisfy $\mathcal{V}_i$.


\subsection{Dataset Construction}
\vspace{-1mm}
As shown in Figure~\ref{fig:dsworld_pipeline}, \DSAgentBench{} was built through a rigorously controlled, three-stage process: 
\Ni sourcing datasets from diverse real-world platforms, \Nii collaborative task and evaluator design, and \Niii dual-annotator verification, where each task was independently executed and validated for accuracy and reproducibility.

\vspace{-1mm}
\subsubsection{Dataset Sourcing} 
\vspace{-1mm}
We collected data from diverse real-world sources to ensure agents encounter the variety of formats typical in practical data science. We primarily collected structured tabular data, supplemented by image and text data, and exhibit varying structural complexity, ranging from flat single-table and multi-table CSVs to normalized relational schemas requiring multi-table joins. These datasets are drawn from Kaggle competition files (prioritizing top-downloaded and highly rated examples), popular OpenML datasets used in academic evaluation, SQLite databases reflecting production relational storage, GitHub repositories such as Plotly datasets, and web-accessible data retrieved through API.

\vspace{-1mm}
\subsubsection{Task and Evaluator Design}
\vspace{-1mm}
To ensure that \DSAgentBench{} reflects real-world data science practice, we grounded its design in both prior literature and empirical analysis. Prior studies show that Python-based environments such as Jupyter notebooks, VS Code, spreadsheets, GitHub, and terminals form the core of everyday data science workflows~\cite{zhang2020data}. Building on this foundation, we manually analyzed 100 high-ranking Kaggle notebooks from popular and historically influential competitions to identify recurring analytical workflows and question types. Two annotators with over five years of data-science experience independently derived an initial task taxonomy, which we then expanded using LLMs to surface underrepresented patterns and ensure comprehensive coverage of the 
data-science lifecycle.

Based on this taxonomy, \DSAgentBench{} organizes tasks into six capability categories aligned with key stages of the data-science workflow (Table~\ref{tab:dsworld_updated_lifecycle}). \textbf{Data acquisition} tasks assess an agent’s ability to locate, retrieve, and load data from heterogeneous sources and formats. \textbf{Exploratory data analysis} tasks evaluate initial sense-making through statistical summaries, filtering, and pattern discovery (e.g., outlier detection and correlation analysis). \textbf{Feature engineering} tasks test data transformation and representation, including feature construction, encoding, scaling, and dimensionality reduction. \textbf{Modeling} tasks focus on training and comparing predictive models under realistic constraints such as data imbalance and metric selection. \textbf{Evaluation and deployment} tasks measure statistical rigor via validation strategies, hyperparameter tuning, and significance testing. Finally, \textbf{visualization and reporting} tasks assess an agent’s ability to communicate insights through correctly grounded visualizations, interactive plot inspection, structured analytical reports, and PowerPoint reporting artifacts (see App.~\ref{appendix:task_categories} for details).

Guided by this taxonomy, four expert annotators collaboratively developed 275 tasks over a three-month period (approximately 400 working hours).  Each task includes: (i) a natural language instruction, (ii) an executable environment configuration, and (iii) a deterministic Python evaluation function for automatic verification.
Tasks were authored by experts using a curated taxonomy; Vision-language models (GPT-5, Claude 4.5 Sonnet, and Gemini 3) were used only to refine wording and identify edge cases, while all task logic, expected outputs, and evaluators were defined and validated by humans. This human-in-the-loop process ensures that tasks reflect realistic analytical reasoning rather than automatically generated QA pairs. Evaluators score semantic correctness of the final outputs (not just successful code execution), enabling rigorous, execution-based benchmarking.


\subsubsection{Dual Annotator Verification}
\vspace{-1mm}
Each task underwent structured dual review for quality assurance. One human annotator (the ``creator") developed the task specification and evaluation logic, while a second annotator (the "verifier") independently assessed instruction clarity, verified executability by running the task with a baseline agent, and validated evaluator correctness. Initial agreement on task quality defined as independent approval by both annotators that the task and evaluator were correct and required no revision was 86\%, with the remaining tasks revised through iterative discussion to refine instructions, fix evaluation logic, or adjust environment configurations. 
All 275 tasks ultimately achieved mutual approval, ensuring every benchmark entry is reproducible, technically correct, and fully executable.

\begin{figure}[t!]
    \centering
    \includegraphics[width=.85
    \columnwidth]{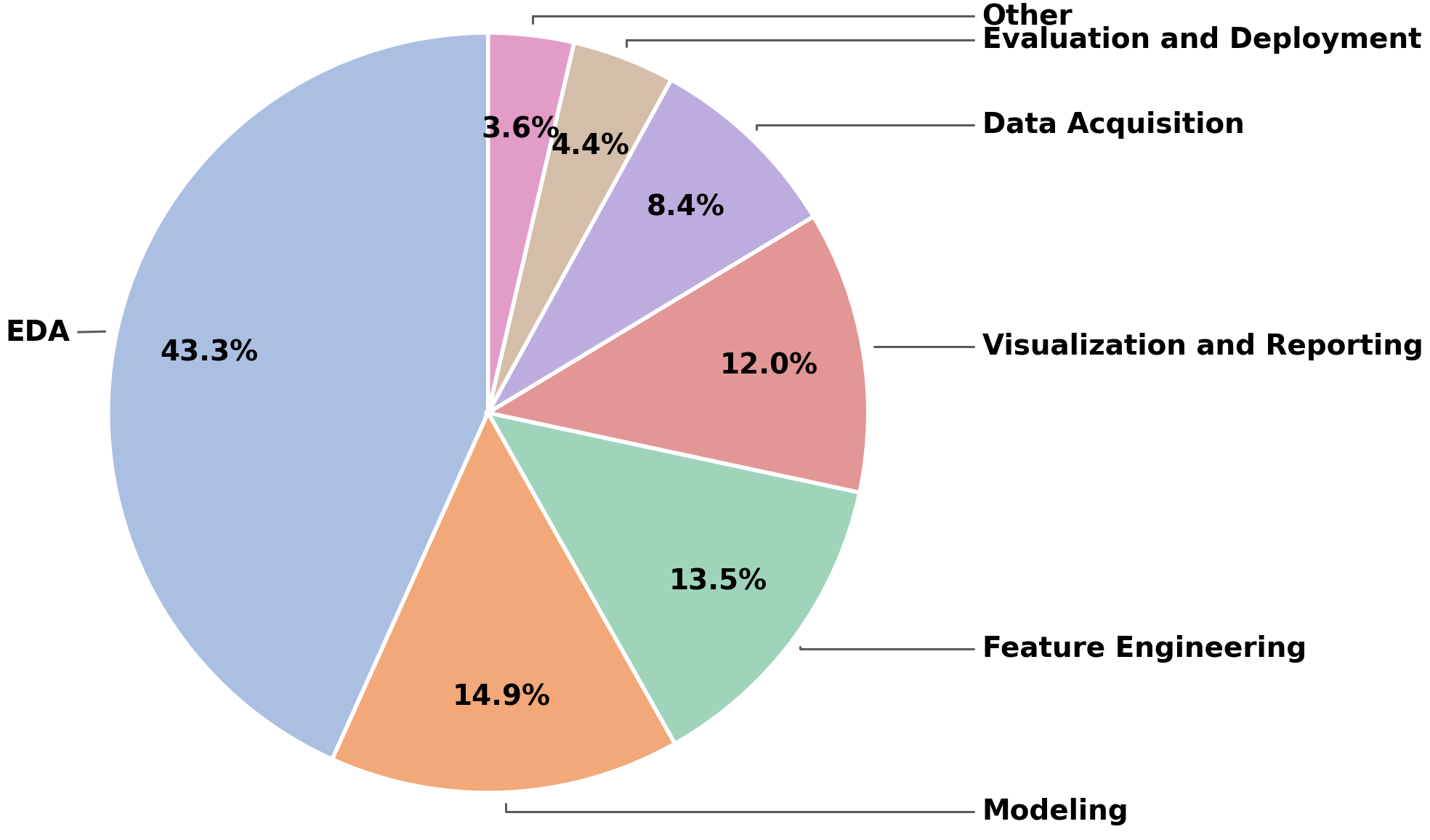}
\caption{Task category distribution in our benchmark, reflecting real-world data science practice.} 
    \label{fig:dsworld-task-categories}
\end{figure}







\subsection{Dataset Diversity \& Statistics}
\DSAgentBench{} is designed to stress-test agent robustness under realistic data-science conditions by varying task structure, difficulty, data modality, source, and tool requirements (Table~\ref{tab:dsworld-composition}). Task categories are well distributed across the data-science lifecycle (Figure~\ref{fig:dsworld-task-categories}), with exploratory data analysis (EDA) comprising the largest share (43.3\%). This distribution aligns with industry observations that data scientists spend 40-60\% of their time on data preparation and exploration~\citep{anaconda2025state}.

The benchmark emphasizes challenging, long-horizon reasoning. Nearly half of the tasks are hard (47.6\%), with 46.9\% medium and only 5.5\% easy. 
Difficulty levels are assigned based on multiple factors, including (i) the number of required steps (easy: 1–2; medium: 3–4; hard: 5+), (ii) analytical sophistication (e.g., simple filtering versus model building and comparison), and (iii) tool coordination requirements (single versus multiple tools). Two expert annotators independently labeled task difficulty, resolving disagreements through discussion. 
Moreover, 56.7\% of tasks involve multi-stage workflows—such as load–transform–visualize or model–evaluate–refine—requiring agents to iteratively execute code, inspect intermediate outputs, and adapt later actions, mirroring real-world data science, with tasks averaging 4–5 analytical steps.

Data and source diversity further increase task complexity. While tabular data dominates (95.3\%), reflecting real-world structured workflows, the benchmark also includes image (3.6\%) and text (1.1\%) components, introducing multimodal reasoning challenges; task design remains modality-agnostic.
Datasets are drawn from a wide range of real-world platforms, including GitHub (37.1\%), Kaggle (29.8\%), OpenML (18.9\%), and SQLite databases (7.6\%). 
 Finally, tasks run in real operating systems and require coordinated use of multiple tools—typically Python alongside VS Code, Jupyter Notebook, and Chrome—often involving three or more tools per task. This moves evaluation beyond static code generation to realistic workflow execution and cross-application reasoning.

\definecolor{rowlight1}{RGB}{245,245,245}
\definecolor{rowlight2}{RGB}{235,245,255}
\definecolor{rowlight3}{RGB}{255,240,225}
\definecolor{rowlight4}{RGB}{240,255,240}
\definecolor{rowlight5}{RGB}{255,250,230}

\begin{table}[t!]
\centering
\caption{Key characteristics of \DSAgentBench{}.}


\footnotesize
\setlength{\tabcolsep}{6pt}

\scalebox{0.85}{%
\begin{tabular}{l|p{0.75\columnwidth}}
\toprule
\textbf{Dimension} & \textbf{Distribution Breakdown} \\
\midrule

\rowcolor{rowlight1}
\textbf{Complexity} 

& Hard: 47.6\% \;|\; Medium: 46.9\% \;|\; Easy: 5.5\% \\


\rowcolor{rowlight2}

\textbf{Stage Type} 
& Multi-Stage: 56.7\% \;|\; Single-Stage: 43.3\% \\


\rowcolor{rowlight3}
\textbf{Modality} 
& Tabular: 95.3\% \;|\; Image: 3.6\% \;|\; Text: 1.1\% \\


\rowcolor{rowlight4}
\textbf{Source}

& GitHub: 37.1\% \;|\; Kaggle: 29.8\% \;|\; OpenML: 18.9\% \;|\; SQLite: 7.6\% \;|\; Web: 6.5\% \\


\rowcolor{rowlight5}
\textbf{Tools Used} 
& Python: 100.0\% \;|\; VS Code: 81.1\% \;|\; Jupyter Notebook: 18.9\% \;|\; Chrome: 10.2\% \\
\bottomrule
\end{tabular}
}
\label{tab:dsworld-composition}
\vspace{-3mm}
\end{table}

%% file: emnlp2020-templates/experiment.tex
\definecolor{human_baseline}{RGB}{255, 245, 170}
\definecolor{open_models_below_4B}{RGB}{185, 235, 255}
\definecolor{open_models_7B_12B}{RGB}{255, 219, 187}
\definecolor{closed_models}{RGB}{240, 240, 240}
\definecolor{chart_specific_models}{RGB}{217, 240, 211}

\definecolor{human_baseline}{HTML}{E8F5E9}

\begin{table*}[t]
\centering
\caption{
DSAgentBench accuracy (\%) under two observation settings (Screenshot and Screenshot + A11y Tree) across data-science lifecycle tasks: Data Acquisition (DA), Exploratory Data Analysis (EDA), Feature Engineering (FE), Modeling, Visualization (Vis), and Evaluation (Eval). Results for open-source models are reported in Table~\ref{tab:dsagentbench-open}.
}

\resizebox{\textwidth}{!}{%
\scriptsize
\begin{tabular}{l|ccccccc|ccccccc}

\toprule
\multirow{2}{*}{\textbf{Model}}
& \multicolumn{7}{c|}{\textbf{Screenshot}}
& \multicolumn{7}{c}{\textbf{Screenshot + A11y Tree}} \\

\cmidrule(lr){2-8} \cmidrule(lr){9-15}

& \textbf{DA}
& \textbf{EDA}
& \textbf{FE}
& \textbf{Model}
& \textbf{Vis}
& \textbf{Eval}
& \textbf{Overall}

& \textbf{DA}
& \textbf{EDA}
& \textbf{FE}
& \textbf{Model}
& \textbf{Vis}
& \textbf{Eval}
& \textbf{Overall} \\

\midrule


\rowcolor{closed_models!50}
GPT-4o
& 0.00 & 27.06 & 24.24 & 16.67 & 7.14 & 33.33 & 19.34
& 13.04 & 29.66 & 35.14 & 21.95 & 15.62 & 8.33 & 24.54 \\
\rowcolor{closed_models!50}
O4-mini
& 4.35 & 1.68 & 2.70 & 0.00 & 3.03 & 0.00 & 1.82
& 0.00 & 4.20 & 5.41 & 0.00 & 0.00 & 0.00 & 2.55 \\

\rowcolor{closed_models!50}
GPT-5-mini
& 8.70 &17.28 & 25.00 & 3.57 & 21.43 & 0.00 & 15.20
& 4.55 & 24.14 & 27.03 & 7.50 & 21.88 & 0.00 & 19.03 \\

\rowcolor{closed_models!50}
GPT-5
& 17.39 & 27.41 &  27.03 & 26.83  & 9.09 & 16.67 & 23.63
& 26.08 & 33.44  & 29.72 &  32.22 & 15.15 &33.33  & 29.81 \\

\rowcolor{closed_models!50}
Gemini-2.5-Pro
& 4.55 & 17.86 & 16.13 & 17.24 & 10.71 & 16.67 & 14.49
& 4.35 & 22.78 & 33.33 & 14.81 & 30.77 & 0.00 & 20.81 \\

\rowcolor{closed_models!50}
OpenAI CUA
& 0.00 & 10.59 & 12.50 & 0.00 & 14.29 & 0.00 & 8.13
& 4.35 & 9.52 & 6.06 & 2.86 & 6.67 & 0.00 & 6.61 \\

\rowcolor{closed_models!50}
Claude-4-Sonnet
& 0.00 & 7.46 & 4.17 & 0.00 & 3.85 & 0.00 & 4.55
& 0.00 & 4.86 & 0.00 & 1.00 & 11.11 & 0.00 & 4.64 \\

\rowcolor{closed_models!50}
Claude-4.5-Sonnet
&4.76  & 4.17 & 0.00 & 3.45 & 3.23& 0.00 & 3.17
& 8.33 & 12.31 &12.50 & 4.55 & 4.17 & 0.00 & 9.21 \\

\rowcolor{closed_models!50}
Claude-4.6-Sonnet
& \textbf{43.48}& \textbf{57.98} & \textbf{51.35}& \textbf{41.46}& \textbf{39.39} &\textbf{50.00} &\textbf{50.55}
& \textbf{47.82} & \textbf{64.88} & \textbf{56.75} & \textbf{46.34} & \textbf{42.42}  & \textbf{66.67}  & \textbf{56.70} \\

\rowcolor{human_baseline}
\textbf{Human Performance}
&73.91  & 90.76 & 91.89 & 73.17 &78.79  &83.33 & 85.09
&73.91  & 90.76 & 91.89 & 73.17 &78.79  &83.33 & 85.09 \\







\bottomrule
\end{tabular}
\vspace{-3mm}
}
\label{tab:dsagentbench-lifecycle}
\vspace{-3mm}
\end{table*}

\begin{table*}[t]
\centering
\caption{DSAgentBench \textbf{accuracy} (\%) across stage type, task complexity, and tool usage under different settings.}
\vspace{-3mm}
\resizebox{\textwidth}{!}{%
\begin{tabular}{l|cc|ccc|cc|cc|ccc|cc}
\toprule
\multirow{2}{*}{\textbf{Model}}
& \multicolumn{7}{c|}{\textbf{Screenshot}}
& \multicolumn{7}{c}{\textbf{Screenshot + A11y Tree}} \\

\cmidrule(lr){2-8}
\cmidrule(lr){9-15}

& \textbf{Single}
& \textbf{Multi}
& \textbf{Easy}
& \textbf{Medium}
& \textbf{Hard}
& \textbf{VS Code}
& \textbf{Jupyter}

& \textbf{Single}
& \textbf{Multi}
& \textbf{Easy}
& \textbf{Medium}
& \textbf{Hard}
& \textbf{VS Code}
& \textbf{Jupyter} \\

\midrule



\rowcolor{closed_models!50}
GPT-4o
& 26.97 & 13.52 & 38.46 & 24.74 & 11.79 & 14.69 & \textbf{39.22}
& 35.59 & 16.13 & 46.67 & 34.38 & 12.31 & 23.87 & 27.45 \\

\rowcolor{closed_models!50}
O4-mini
& 4.20 & 0.00 & 20.00 & 1.55 & 0.00 & 1.79 & 1.92
& 5.04 & 0.64 & 13.33 & 3.88 & 0.00 & 2.69 & 1.92 \\

\rowcolor{closed_models!50}
GPT-5-mini
& 24.71 & 8.00 & 38.46 & 24.47 & 9.41 & 15.23 & 11.00
& 31.03 & 9.87 & 66.67 & 27.20 & 5.47 & 14.68 & 38.00 \\

\rowcolor{closed_models!50}
GPT-5
&27.73  & 20.51 & 53.33 & 27.13 & 16.79 &25.10  & 17.30
&39.49  &22.43  &66.67 & 32.55 & 22.89 & 27.35 &40.38  \\

\rowcolor{closed_models!50}
Gemini-2.5-Pro
& 19.12 & 10.92 & 46.15 & 15.62 & 9.18 & 17.31 & 2.40
& 32.10 & 12.93 & 53.85 & 28.74 & 9.28 & 17.45 & 31.25 \\

\rowcolor{closed_models!50}
OpenAI CUA
& 13.64 & 4.00 & 23.08 & 13.40 & 1.20 & 9.49 & 2.30
& 10.68 & 3.60 & 15.38 & 10.43 & 1.84 & 5.21 & 12.60 \\

\rowcolor{closed_models!50}
Claude-4-Sonnet
& 6.67 & 2.93 & 8.33 & 4.65 & 4.00 & 5.60 & 0.00
& 8.06 & 2.10 & 25.00 & 6.33 & 0.64 & 4.86 & 3.70 \\

\rowcolor{closed_models!50}
Claude-4.5-Sonnet
& 3.16 & 3.17 & 16.67 & 3.51 & 1.05 & 4.14 & 0.00
& 14.49 & 4.82 &36.36 & 10.00 & 3.28 & 8.74 & 10.20 \\

\rowcolor{closed_models!50}
Claude-4.6-Sonnet
&\textbf{59.66}  & \textbf{43.60} &\textbf{73.33}  & \textbf{55.04} &\textbf{43.52}  &\textbf{54.01}  &35.71
&\textbf{63.86}  &\textbf{51.24}  & \textbf{86.68} & \textbf{60.62} & \textbf{49.41} &\textbf{56.92}  & \textbf{55.77} \\

\bottomrule
\end{tabular}
}
\label{tab:dsagentbench-stage-complexity-tool-obs}
\vspace{-3mm}
\end{table*}

\vspace{-1mm}
\subsection{Models}
\vspace{-1mm}

We evaluate a diverse set of vision-language agents spanning closed-source, hybrid, and open-source architectures. Closed-source agents include GPT-4o~\cite{openai2024gpt4technicalreport}, GPT-5-mini, GPT-5 ~\cite{openai2025gpt5}, O4-mini~\cite{openai2024gpt4technicalreport}, Claude Sonnet 4, 4.5 and 4.6, Gemini 2.5 Pro~\cite{geminipro25}, and OpenAI's Computer Agent~\cite{openaiComputerUsingAgent}, all of which adopt unified architectures where a single vision-language model handles both UI grounding and action planning. Hybrid models combine open-source grounding models with proprietary planning models, including Jedi-3B and Jedi-7B paired with GPT-4o~\cite{jedi}. Open-source models include UI-TARS 2B and UI-TARS 7B~\cite{uitars}, GUI-OWL-7B~\cite{ye2025mobileagentv3fundamentalagentsgui}, and OpenCUA-72B \cite{wang2025opencua} representing fully open architectures trained specifically for GUI interaction.


\subsection{Execution-Based Evaluation and Metrics}
Each task
is paired with a deterministic Python evaluator executed after the agent completes its interaction sequence. 
After task termination, the evaluator collects relevant artifacts from the virtual machine, including scripts, generated files, visualizations, or trained models, and scores them against task-specific evaluation criteria defined for that task. For example, in a \textit{correlation analysis task}, the evaluator verifies that the required output file exists, loads the computed Pearson correlation coefficient, and checks whether it matches the expected value within a tolerance of $\epsilon = 0.01$. For \textit{visualization tasks}, the evaluator checks that a visualization is generated and verifies required metadata such as labeled axes, titles, legends, and correct data mappings. In addition, GPT-4o is used as a visual judge to assess visualization quality and semantic alignment with the task requirements. To avoid 
evaluation circularity, we use Gemini-2.5-Pro as the visual judge 
for all GPT-4o generated outputs, while GPT-4o serves as the judge 
for outputs from other models. Only $\sim$10\% of tasks invoke an LLM-based judge, 
and only after deterministic validation gates. 
For \textit{classification tasks}, evaluators verify that trained models meet minimum performance thresholds (e.g., accuracy or F1 $\geq 0.7$). 
Each evaluator produces a continuous score in $[0,1]$.

We use \textit{task success rate} as the primary evaluation metric, defined as the percentage of tasks achieving an overall score greater than or equal to 0.95. This threshold allows minor tolerance for numerical precision or formatting variation while ensuring semantic correctness of the analytical outcome. We also report the \textit{average evaluation score} across all tasks to provide a  detailed assessment of agent performance. Representative evaluator designs are provided in ~\Cref{app:evaluation} (Tables~\ref{tab:numerical-eval-structure} and~\ref{tab:visual-eval-structure}).





%% file: emnlp2020-templates/Evaluation.tex
\vspace{-1mm}
\subsection{Main Results}

Table~\ref{tab:dsagentbench-lifecycle} presents the performance of 
agents across both observation settings: \textit{Screenshot-only} and the hybrid \textit{Screenshot + A11y Tree} configuration. As human reference points, three participants, two applied scientists and one master’s graduate, achieved an overall success rate of 85.09\% under the same task environment and deterministic evaluation protocol. All results use the fixed evaluation setup described in App.~\ref{sec:impl-details}. The results reveal a clear performance gap between current agents and the demands of real-world end-to-end data science workflows. Claude-4.6-Sonnet achieves the strongest agent performance, reaching 56.70\% overall accuracy under the Screenshot + A11y Tree setting, followed by GPT-5 at 29.81\%. Other closed-source models, including GPT-4o, Gemini-2.5-Pro, and GPT-5-mini, achieve lower overall accuracy of around 20\%, highlighting persistent long-horizon execution challenges. In contrast, open-source agents achieve at most 1\% accuracy under the Screenshot-only setting, and none support A11y Tree input (Table \ref{tab:dsagentbench-open}). Despite generating
code and issue interface actions, these models consistently fail to ground instructions in the UI state and complete end-to-end workflows.

Performance varies substantially across task types (Table \ref{tab:dsagentbench-lifecycle}). Data acquisition, model validation, and evaluation tasks remain most challenging, reflecting the difficulty of long-horizon reasoning and tool coordination.
Adding accessibility information to screenshots generally improves performance, indicating that structured UI metadata aids grounding and interaction (Tables~\ref{tab:dsagentbench-lifecycle} and~\ref{tab:dsagentbench-stage-complexity-tool-obs}). However, gains vary widely across models, and some struggle to effectively exploit A11y signals. Even under the hybrid setting, overall success rates remain low, underscoring that current agents are still far from autonomous data science workflows.

\definecolor{open_models_below_4B}{RGB}{185, 235, 255}
\definecolor{open_models_7B_12B}{RGB}{255, 219, 187}
\definecolor{easy_color}{RGB}{210, 242, 204}
\definecolor{medium_color}{RGB}{255, 236, 179}
\definecolor{hard_color}{RGB}{255, 204, 188}
\definecolor{overall_color}{RGB}{204, 229, 255}

\subsection{Ablation Studies}

To diagnose 
performance trends, we analyze variations across task complexities, workflow structure, and tool usage (Table~\ref{tab:dsagentbench-stage-complexity-tool-obs}). Tasks executed in Jupyter consistently outperform those using VS Code, largely due to fewer terminal and environment-related failures. Single-stage tasks achieve substantially higher success rates than multi-stage workflows, highlighting the difficulty of maintaining state, recovering from errors, and coordinating tools across longer execution chains. Performance also degrades monotonically with task difficulty, with hard tasks remaining the most challenging due to long-horizon reasoning and iterative refinement.


We additionally ablate the maximum interaction budget by evaluating step limits of 15, 30, and 50 (Table~\ref{tab:step_ablation}). Increasing the budget from 15 to 50 steps yields only marginal gains in task success rate (24.54\% $\rightarrow$ 25.81\%) and average score (0.55 $\rightarrow$ 0.57). The small gain suggests that performance is not primarily limited by the action budget; instead, failures reflect a combination of grounding, planning, reasoning, tool orchestration, and analytical execution errors. We also run a terminal-first ablation with GPT-4o in the screenshot-only setting, prompting the agent to prefer terminal or command-line workflows whenever they could solve the current subtask more directly and reliably. Relative to the standard screenshot-only setting, performance changes only marginally, from 19.34\% to 20.73\%, suggesting that terminal-first execution alone does not substantially reduce the benchmark difficulty.

\section{Error Analysis}
\label{error_an}
To better understand error patterns, we conduct a targeted analysis of 604 manually inspected runs from closed-source models using Screenshot + A11y Tree observations, along with 150 sampled runs from open-source models (Screenshot-only).

\noindent\textbf{Root Causes of Failure.} (Table~\ref{tab:root-cause-failure}) shows that DSAgentBench evaluates both environment grounding and data-science execution. Open-source agents fail almost entirely due to grounding errors (97–98\%), revealing poor alignment with desktop state. Stronger closed-source agents show mixed failures: GPT-4o still fails mainly from grounding, while better-grounded models (Gemini-2.5-Pro, GPT-5, Claude-4.6-Sonnet) exhibit terminal, code, and reasoning failures. This indicates success requires both robust UI control and sound data-science reasoning in real environments.

\noindent\textbf{Temporal Failure Structure and Recovery.} Table~\ref{tab:trajectory-analysis} analyzes when failures occur using the first failure step (FF), defined as the earliest step where an error is detected. CUA and GUI-OWL-7B exhibit predominantly late-stage failures, with over 93\% occurring after prolonged interaction, reflecting extended but ineffective exploration due to poor grounding. In contrast, GPT-4o and Gemini-2.5-Pro show a higher proportion of early and mid-trajectory failures, indicating weaker robustness in initial grounding and planning. Open-source models fail especially early, often unable to correctly open or control the terminal. Budget exhaustion further differentiates behavior: CUA and open-source agents almost always exhaust their step budgets, while Gemini 2.5 Pro and GPT-5-Mini terminate earlier on a substantial fraction of tasks.

\noindent\textbf{Efficiency of Successful Runs.}
As shown in Table~\ref{tab:efficiency-success}, Gemini-2.5-Pro completes tasks most efficiently (6.76 steps on average), followed by GPT-5-Mini (7.33), GPT-4o (10.03), and CUA (15.00). This reveals a clear trade-off between exploration depth and execution efficiency: models that act conservatively and terminate earlier achieve faster successes but also incur higher early-failure rates.

%% file: emnlp2020-templates/Appendix.tex
\appendix
\section{Appendices}

\subsection{Task Categories}
\label{appendix:task_categories}

\begin{table}[h]
\scriptsize
\centering
\caption{Lifecycle grouping based on DS-World task category hierarchy.}
\label{tab:dsworld_updated_lifecycle}
\resizebox{\textwidth}{!}{%
\begin{tabular}{l|l}
\midrule
\textbf{Category} & \textbf{Lifecycle / Macro Group} \\
\midrule

\rowcolor[HTML]{CFE2F3}
Data Loading and Multi-Table Joining & Data Acquisition \\
\rowcolor[HTML]{CFE2F3}
Web Data Scraping & Data Acquisition \\

\midrule

\rowcolor[HTML]{E2F0CB}
Summary Statistics & EDA \\
\rowcolor[HTML]{E2F0CB}
Data Cleaning and Missing-Value Handling & EDA \\
\rowcolor[HTML]{E2F0CB}
Data Filtering and Conditional Queries & EDA \\
\rowcolor[HTML]{E2F0CB}
Correlation and Relationship Analysis & EDA \\
\rowcolor[HTML]{E2F0CB}
Grouping and Aggregation & EDA \\
\rowcolor[HTML]{E2F0CB}
Outlier Detection & EDA \\

\midrule

\rowcolor[HTML]{FFF3C8}
Feature Engineering and Transformation & Feature Engineering \\
\rowcolor[HTML]{FFF3C8}
Dimensionality Reduction & Feature Engineering \\
\rowcolor[HTML]{FFF3C8}
Standardization and Normalization & Feature Engineering \\
\rowcolor[HTML]{FFF3C8}
Feature Importance and Interpretability & Feature Engineering \\

\midrule

\rowcolor[HTML]{F9E0E0}
Classification & Modeling \\
\rowcolor[HTML]{F9E0E0}
Regression & Modeling \\
\rowcolor[HTML]{F9E0E0}
Clustering & Modeling \\
\rowcolor[HTML]{F9E0E0}
Ensemble Methods & Modeling \\
\rowcolor[HTML]{F9E0E0}
Imbalanced Data Handling & Modeling \\

\midrule

\rowcolor[HTML]{E7D7FF}
Model Validation and Evaluation & Evaluation and Deployment \\
\rowcolor[HTML]{E7D7FF}
Hyperparameter Tuning & Evaluation and Deployment \\
\rowcolor[HTML]{E7D7FF}
Statistical Testing & Evaluation and Deployment \\

\midrule

\rowcolor[HTML]{D4E8FF}
Visualization and Chart Generation & Visualization and Reporting \\
\rowcolor[HTML]{D4E8FF}
Data Export and Reporting & Visualization and Reporting \\

\midrule
\end{tabular}
}
\end{table}
\paragraph{Data Acquisition}
These tasks measure an agent’s ability to locate, retrieve, and load data from diverse sources, including CSV files, multi-table joins, URL-based datasets, SQLite databases, and web-scraped content. Success depends on correctly identifying file paths and handling retrieval failures, reflecting real-world challenges in the earliest stage of analysis. 

\definecolor{rowlight1}{RGB}{245,245,245}
\definecolor{rowlight2}{RGB}{235,245,255}
\definecolor{rowlight3}{RGB}{255,240,225}
\definecolor{rowlight4}{RGB}{240,255,240}
\definecolor{rowlight5}{RGB}{255,250,230}
\definecolor{rowlight6}{RGB}{245,235,255}

\begin{table*}[t]
\centering
\caption{Representative simplified Example Tasks in \DSAgentBench{} by Category}
\label{tab:dsworld_task_cat}

\resizebox{\textwidth}{!}{%
\begin{tabular}{l|c|p{0.70\textwidth}}
\toprule
\textbf{Task Category} & \textbf{\# Tasks} & \textbf{Example Task} \\
\midrule

\rowcolor{rowlight1}
\textbf{Data Acquisition} 
& 23
& Open the database, inspect the available tables and relevant columns, identify the customer ID and transaction fields, merge the required records, compute total and average transaction amounts per customer, calculate a risk score, and save the aggregated results to an output file. \\

\rowcolor{rowlight2}
\textbf{Exploratory Data Analysis} 
& 119
& Inspect a SQLite database schema, identify meal- and workout-related fields, join the relevant tables, compute an efficiency index, rank workout categories, and save the top five results to a CSV file. \\

\rowcolor{rowlight3}
\textbf{Feature Engineering} 
& 37
& Load a diabetes dataset, create a new insulin resistance feature, handle missing or zero values using median imputation, compute correlation with insulin levels, group results by outcome, and save summary statistics to output files. \\

\rowcolor{rowlight4}
\textbf{Modeling} 
& 41
& Load and merge multiple retail datasets by store and date, train multiple regression models to predict weekly sales, stack them using a meta-learner, perform cross-validation, and save the average error metric to a summary file. \\

\rowcolor{rowlight5}
\textbf{Evaluation and Deployment} 
& 12
& Perform multi-stage hyperparameter tuning for a gradient boosting model using cross-validation, select the best configuration based on error metrics, and save the final model parameters and performance results. \\

\rowcolor{rowlight6}
\textbf{Visualization and Reporting} 
& 33
& Analyze the relationship between body mass index, glucose level, and age with diabetes outcome, generate and inspect interactive visualizations, and create a PowerPoint report summarizing the key findings with appropriate charts, labels, and legends.. \\

\bottomrule
\end{tabular}
}
\end{table*}

\paragraph{Exploratory Data Analysis (EDA)}
EDA tasks evaluate initial sense-making and inspection of raw data through summary statistics, filtering, outlier detection, correlation analysis, and group-wise trend identification. Agents must form meaningful interpretations before modeling, mirroring how human analysts spend a substantial portion of their workflow time.

\paragraph{Feature Engineering}
These tasks assess transformation and representation learning, requiring agents to create new features, scale numeric variables, encode categoricals, reduce dimensionality, and analyze feature importance. Effective feature engineering is essential for converting raw data into predictive structure rather than modeling unprocessed inputs.

\paragraph{Modeling}
Modeling tasks focus on building predictive systems through classification, regression, clustering, ensemble construction, and handling data imbalance. Many tasks require training multiple models and comparing validation metrics, reflecting realistic scenarios where model performance rather than code syntax determines success. 

\paragraph{Evaluation and Deployment}
These tasks measure statistical rigor, model reliability, and interpretability. Examples include hyperparameter tuning, K-fold validation, metric comparison, hypothesis testing, and statistical significance evaluation. Agents must justify output quality using measurable evidence rather than relying solely on execution.

\paragraph{Evaluation and Deployment}
Visualization tasks test an agent’s ability to communicate insights through bar charts, scatter plots, heatmaps, pairplots, and model comparison graphs, as well as structured text or file-based reports. Evaluators verify semantic correctness through labeled axes, titles, legends, and data mappings, ensuring meaningful analysis rather than surface-level image generation. We provide detailed breakdown of lifecycle grouping of task category and task examples in Table~\ref{tab:dsworld_updated_lifecycle} and Table~\ref{tab:dsworld_task_cat}, respectively.

\begin{figure*}[t]
    \includegraphics[width=0.85\textwidth]{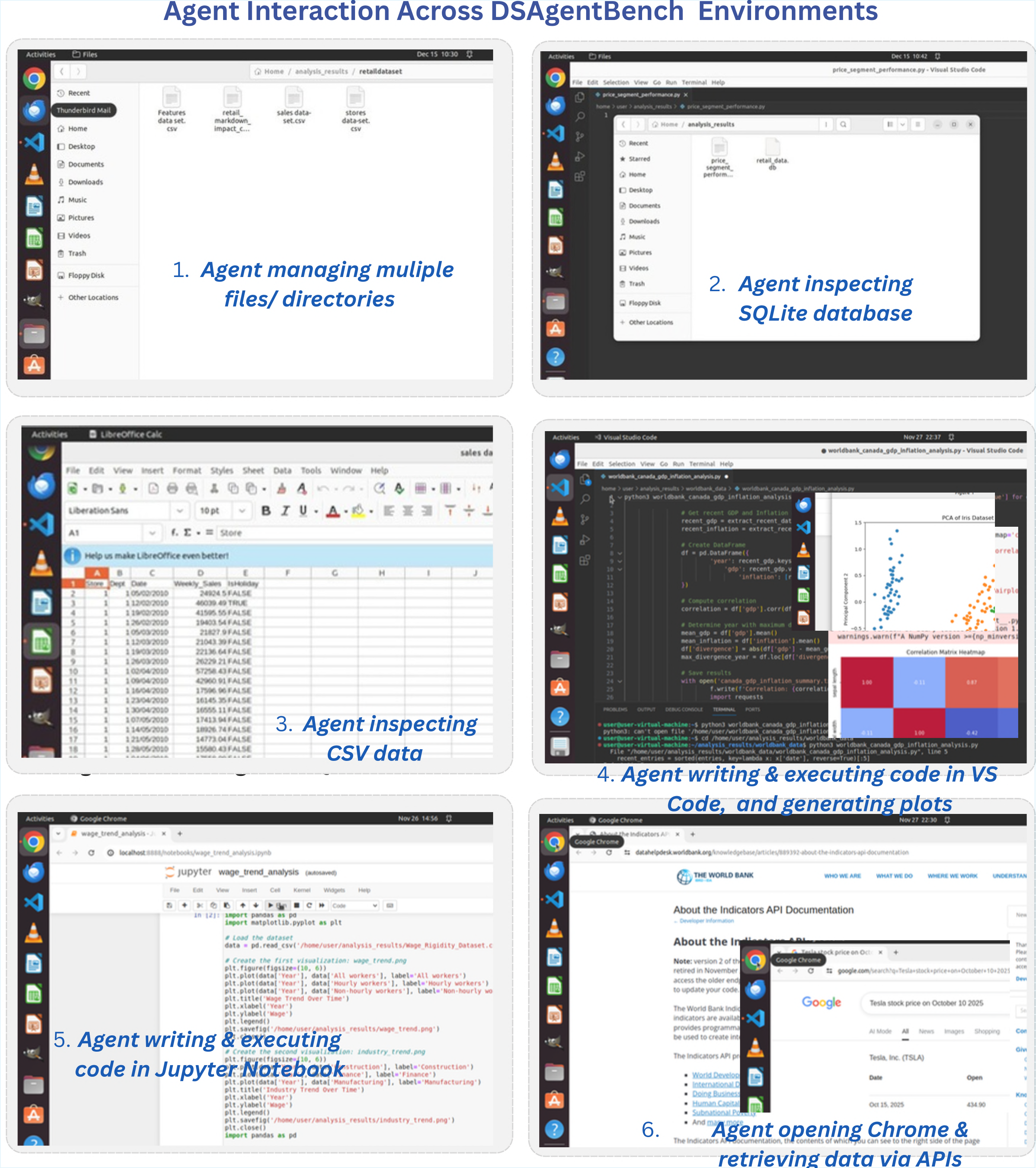}
    \caption{
    Example environments in \DSAgentBench{}: agent interaction across data-science environments, demonstrating autonomous coordination of file systems, databases, notebooks, IDEs, visualization, execution, debugging, and web-based data retrieval.
    }
    \label{fig:agent_interaction}
\end{figure*}
\section{Methodology}
\label{app:method}

\subsection{Environment and Inference Setup}
We use different computational setups for closed-source and open-source models. Closed-source models are accessed exclusively through provider APIs and therefore do not require local model hosting. For these models, we run our experiments on an Ubuntu virtual machine created using VMware, where we deploy the our framework to manage API-based inference and evaluation.

For open-source models, we adopt a self-hosted setup to improve cost efficiency and enable concurrent inference across multiple models. Specifically, we run Docker-based virtual machines on Google Cloud Platform (GCP) instances with GPU support. Each model is served using the vLLM inference engine, and inference is performed by querying the locally hosted vLLM endpoints. Each GCP instance runs \texttt{Ubuntu 22.04 LTS} on an \texttt{n1-standard-4} machine with a $200$ GB persistent standard disk.

\subsection{Initial State Setup}
As our experiments are conducted within the OSWorld environment, we retain its default system configuration, which includes pre-installed applications such as \textbf{VS Code}, \textbf{Google Chrome}, and other standard utilities. To support interaction for the UI-TARS model family, we additionally install the \texttt{pyperclip} library, enabling programmatic copy–paste operations during code execution.

We then prepare a lightweight execution environment for reproducible analysis. Specifically, we install the \textbf{Jupyter Notebook} package to enable notebook-based workflows and create a dedicated directory for storing intermediate and final analysis results. Required input data are downloaded to predefined filesystem locations, and a task-specific Python script is initialized as the main execution entry point. Finally, the analysis environment is launched either through Jupyter Notebook or a code editor, enabling interactive development and execution.

\subsection{Agent Prompt Templates}

This appendix summarizes the canonical system prompts used to control agent behavior under different observation modalities. Rather than listing full implementation-level prompts inline, we abstract each prompt by its observation inputs, output constraints, and grounding assumptions.

\begin{table*}[t]
\centering
\caption{Canonical agent prompt templates used under different observation modalities. One representative prompt is shown per setting; other variants differ only in minor wording or formatting.}
\resizebox{\textwidth}{!}{%
\begin{tabular}{l|l|p{9.5cm}}
\toprule
\rowcolor{gray!30}
\textbf{Observation Modality} & \textbf{Output Format} & \textbf{Prompt Description} \\
\midrule
\rowcolor{gray!5}
Screenshot-only 
& Code / Action 
& The agent receives a full-resolution screenshot of the desktop at each step and predicts grounded mouse and keyboard actions based solely on visual information. Interaction is coordinate-based, without access to image matching or additional screenshots. At each step, the agent must return either executable actions or a control token (\texttt{WAIT}, \texttt{FAIL}, \texttt{DONE}). \\
\rowcolor{gray!5}
Screenshot + A11y Tree 
& Code / Action 
& The agent jointly observes both the screenshot and the accessibility tree extracted via AT-SPI, enabling alignment between visual context and structured UI metadata. This hybrid prompt improves grounding precision and interaction robustness and is used for all main experiments unless stated otherwise. \\
\bottomrule
\end{tabular}
}
\label{tab:prompt-templates}
\end{table*}

For full reproducibility, we will release the complete system prompts, evaluation functions, and model-specific inference scripts for all supported agents as part of the project codebase.

\begin{table}[h]
\centering
\caption{Unified action space listing specific commands and their functions.}
\resizebox{\columnwidth}{!}{%
\begin{tabular}{l|p{9cm}}
\toprule
\rowcolor{gray!30} \textbf{Action Category} & \textbf{Description} \\
\midrule
\rowcolor{gray!5} Mouse Actions & 
Handles spatial interactions including absolute cursor movement (\texttt{MOVE\_TO}, \texttt{DRAG\_TO}), scrolling (\texttt{SCROLL}), and button operations (\texttt{CLICK}, \texttt{RIGHT\_CLICK}, \texttt{DOUBLE\_CLICK}, \texttt{MOUSE\_DOWN}, \texttt{MOUSE\_UP}). \\

\rowcolor{gray!5} Keyboard Actions & 
Manages alphanumeric input via \texttt{TYPING}, discrete key presses (\texttt{PRESS}), simultaneous key combinations (\texttt{HOTKEY}), and raw key state manipulation (\texttt{KEY\_DOWN}, \texttt{KEY\_UP}). \\

\rowcolor{gray!5} Control Actions & 
Special meta-actions to pause execution (\texttt{WAIT}) or terminate the episode by signaling failure (\texttt{FAIL}) or success (\texttt{DONE}). \\

\bottomrule
\end{tabular}
}
\label{tab:action-space}
\end{table}
\subsection{Unified Action Space}

All agents interact with the desktop environment through a unified action space to ensure consistent execution and fair comparison across models and observation modalities (see Tab. \ref{tab:action-space}).





All screen coordinates are defined relative to a fixed resolution of 1920$\times$1080. By constraining agents to this shared action space, performance differences reflect reasoning, grounding, and planning capability rather than interface-specific advantages.

\subsection{Implementation Details}
\label{sec:impl-details}

All agents are evaluated under the unified environment and observation settings described in Section \ref{env} to ensure fair and consistent comparison across models. We use default API configurations for closed-source models, with the temperature fixed at 0.1 to improve determinism while retaining limited response diversity. Closed-source models are evaluated via provider APIs without local hosting, using an OSWorld-based Ubuntu virtual machine running the DSworld framework.

Open-source models are evaluated in a self-hosted setting on Google Cloud Platform (GCP) compute instances using Docker-based virtual machines. Models are served using the vLLM inference engine \cite{vllm}, enabling efficient and concurrent inference across multiple model variants. We will release the complete experimental repository, including prompts, environment configurations, evaluation code, and inference scripts, to facilitate reproducibility.

\paragraph{Key evaluation parameters.}
\begin{itemize}[noitemsep, topsep=0pt, leftmargin=*]
  \item \textbf{Observation:} screenshot, screenshot + accessibility tree 
  \item \textbf{Action space:} \texttt{pyautogui}
  \item \textbf{Resolution:} $1920 \times 1080$
  \item \textbf{Decoding:} temperature $=0.1$, top-$p=0.9$
  \item \textbf{Max output tokens per call:} 2000
  \item \textbf{Task timeout:} 1800 seconds
\end{itemize}


\begin{table}[t]
\centering
\small
\caption{Efficiency of successful runs measured by interaction steps.}
\resizebox{\columnwidth}{!}{
\begin{tabular}{lccc}
\toprule
\rowcolor{gray!30}
\textbf{Model} & \textbf{Mean Steps} & \textbf{Median Steps} & \textbf{Std} \\
\midrule
\rowcolor{closed_models!50} CUA & 15.00 & 15 & 0.00 \\
\rowcolor{closed_models!50} GPT-4.0 & 10.03 & 15 & 7.63 \\
\rowcolor{closed_models!50} Gemini-2.5-Pro & \textbf{6.76} & \textbf{6} & 2.93 \\
\rowcolor{closed_models!50} GPT-5-Mini & 7.33 & 5 & 3.82 \\
\rowcolor{closed_models!50} Claude-Sonnet-4.6 & 10.93 & 12 & 3.04 \\
\bottomrule
\end{tabular}
}
\label{tab:efficiency-success}
\end{table}

\definecolor{open_models}{RGB}{230,245,255}

\definecolor{open_models}{RGB}{230,245,255}  

\begin{table}[t!]
\centering
\scriptsize
\caption{Root-cause failure analysis across models (\%).}
\resizebox{.95\columnwidth}{!}{
\begin{tabular}{lcccc}
\toprule
\textbf{Model} & \textbf{Grounding} & \textbf{Terminal} & \textbf{Code} & \textbf{Logic} \\
\midrule
\rowcolor{closed_models!50} CUA & \textbf{94.44} & 2.78 & 0.46 & 2.31 \\
\rowcolor{closed_models!50} GPT-4.0 & 80.00 & 5.64 & 13.33 & 1.03 \\
\rowcolor{closed_models!50} Gemini-2.5-Pro & 43.08 & \textbf{36.15} & 13.85 & 6.92 \\
\rowcolor{closed_models!50} GPT-5-Mini & 56.92 & 7.69 & 30.77 & 4.62 \\
\rowcolor{closed_models!50} GPT-5 & 41.67 & 8.33 & 38.33 & 11.67 \\

\rowcolor{closed_models!50} Claude-Sonnet-4.5 & 39.32 & 23.07 & 27.60 & 10.00 \\

\rowcolor{closed_models!50} Claude-Sonnet-4.6 & 32.77 & 9.24 & \textbf{43.70} & \textbf{14.29} \\

\midrule
\rowcolor{open_models!50} GUI-OWL-7B & 97.09 & 1.09 & 1.45 & 0.36 \\
\rowcolor{open_models!50} UI-Tars-1.5-7B & \textbf{98.18} & 1.82 & 0.00 & 0.00 \\
\rowcolor{open_models!50} Jedi-7B & 81.89 & 6.69 & 7.09 & 4.33 \\
\bottomrule
\end{tabular}
}
\label{tab:root-cause-failure}
\vspace{-3mm}
\end{table}

\begin{table}[t!]
\centering
\scriptsize
\caption{Ablation results for GPT-4o on \dsworld{} under different interaction step budgets.}
\label{tab:step_ablation}
\resizebox{.90\columnwidth}{!}{%
\begin{tabular}{l|c|c|c}
\toprule
\textbf{Metric} & \textbf{15 Steps} & \textbf{30 Steps} & \textbf{50 Steps} \\
\midrule
\rowcolor{easy_color!50}
Task Success Rate (\%) & 24.54 & 25.45 &25.81  \\
\rowcolor{medium_color!50}
Average Score & 0.55 & 0.56 & 0.57 \\
\bottomrule
\end{tabular}
}
\vspace{-3mm}
\end{table}

\begin{table*}[t]
\centering
\caption{Model-wise performance and trajectory behavior (normalized over 100 tasks per model for comparison). 
FF denotes the \textit{first failure step}, i.e., the earliest step in the trajectory where an error is detected.}
\resizebox{\textwidth}{!}{
\begin{tabular}{lccccccccc}
\toprule
\textbf{Model} & \textbf{\#Tasks} & \textbf{Mean FF Step} & \textbf{Median FF} &
\textbf{Early (\%)} & \textbf{Mid (\%)} & \textbf{Late (\%)} &
\textbf{Budget Exhaustion (\%)} & \textbf{Recovery (\%)} \\
\midrule
\rowcolor{closed_models!50} CUA & 65 & 13.62 & 15 & 3.41 & 3.41 & \textbf{93.17} & \textbf{100.0} & 0.0 \\
\rowcolor{closed_models!50} GPT-4.0 & 100 & 8.77 & 15 & 41.54 & 3.59 & 54.87 & 93.85 & 1.54 \\
\rowcolor{closed_models!50} Gemini-2.5-Pro & 100 & 5.80 & 6 & 42.02 & \textbf{45.38} & 12.61 & 62.31 & \textbf{19.23} \\
\rowcolor{closed_models!50} GPT-5-Mini & 100 & 5.92 & 4 & \textbf{57.38} & 19.67 & 22.95 & 63.08 & 3.08 \\
\midrule
\rowcolor{open_models!50} GUI-OWL-7B & 50 & 14.16 & 15 & 2.91 & 4.00 & 93.09 & 100.0 & 0.0 \\
\rowcolor{open_models!50} UI-Tars-1.5-7B & 50 & 14.40 & 15 & 0.00 & 0.00 & \textbf{100.00} & 100.0 & 0.73 \\
\rowcolor{open_models!50} Jedi-7B & 50 & 12.10 & 15 & 11.02 & 18.11 & 70.87 & 98.43 & 1.18 \\
\bottomrule
\end{tabular}
}
\label{tab:trajectory-analysis}
\end{table*}

\begin{table*}[t]
\centering
\caption{
\DSAgentBench{} \textbf{accuracy} (\%) across data-science lifecycle task types and overall performance under Screenshot settings (\textbf{Open-Source Models}).
}
\resizebox{\textwidth}{!}{%
\begin{tabular}{l|ccccccc|ccccccc}
\toprule
\multirow{2}{*}{\textbf{Model}}
& \multicolumn{7}{c|}{\textbf{Screenshot}}
& \multicolumn{7}{c}{\textbf{Screenshot + A11y Tree}} \\

\cmidrule(lr){2-8} \cmidrule(lr){9-15}

& \textbf{DA}
& \textbf{EDA}
& \textbf{FE}
& \textbf{Model}
& \textbf{Vis}
& \textbf{Eval}
& \textbf{Overall}

& \textbf{DA}
& \textbf{EDA}
& \textbf{FE}
& \textbf{Model}
& \textbf{Vis}
& \textbf{Eval}
& \textbf{Overall} \\

\midrule

\rowcolor{open_models_below_4B!50}
Jedi-3B w/GPT4o
& 0.00 & 0.00 & 0.00 & 0.00 & 0.00 & 0.00 & 0.00
& N/A & N/A & N/A & N/A & N/A & N/A & N/A \\

\rowcolor{open_models_below_4B!50}
Jedi-7B w/GPT4o
& 0.84 & 0.00 & 0.00 & 0.00 & 1.00 & 0.00 & 0.73
& N/A & N/A & N/A & N/A & N/A & N/A & N/A \\

\rowcolor{open_models_7B_12B!50}
UI-Tars-2B
& 0.00 & 0.00 & 0.00 & 0.00 & 0.00 & 0.00 & 0.00
& N/A & N/A & N/A & N/A & N/A & N/A & N/A \\

\rowcolor{open_models_7B_12B!50}
UI-Tars-1.5-7B
& 0.00 & 0.00 & 0.00 & 0.00 & 0.00 & 0.00 & 0.00
& N/A & N/A & N/A & N/A & N/A & N/A & N/A \\

\rowcolor{open_models_7B_12B!50}
GUI-OWL-7B
& 0.00 & 0.00 & 0.00 & 0.00 & 0.00 & 0.00 & 0.00
& N/A & N/A & N/A & N/A & N/A & N/A & N/A \\

\rowcolor{open_models_7B_12B!50}
OpenCUA-72B
& 4.35 & 0.91 & 0.00 & 0.00 & 0.00 & 0.00 & 0.73
& N/A & N/A & N/A & N/A & N/A & N/A & N/A \\

\bottomrule
\end{tabular}
}
\label{tab:dsagentbench-open}
\end{table*}

\subsection{Reproducibility and Environment Portability}
\label{sec:reprod}

DSAgentBench is designed to support reproducible execution while remaining extensible to other desktop environments. Although all experiments in this paper are reported on an Ubuntu-based environment, DSAgentBench is built by extending OSWorld, which supports major desktop operating systems, including Ubuntu/Linux, Windows, and macOS. Each benchmark task specifies the required datasets, file-system state, applications, libraries, setup procedures, and deterministic evaluation metrics. To avoid dependence on changing external data availability, all datasets collected from public sources such as Kaggle, UCI/OpenML, GitHub, and SQLite repositories are pre-downloaded and will be released together with the benchmark.  This task-level configuration makes the benchmark reproducible and portable: the same tasks can be instantiated on other OSWorld-supported operating systems by preparing the corresponding software stack, while reusing the same deterministic evaluators to verify final outputs. In addition, the environment design is extensible to other data-science tools, such as R/RStudio, PyCharm, and cloud-based environments, through additional environment configurations and task-specific setup scripts.

For the small subset of visualization/reporting tasks that require qualitative assessment, we use an LLM-based visual judge only after deterministic validation gates have passed. Rule-based evaluators first verify execution correctness, required output files, data mappings, labels, legends, and task-specific constraints. The LLM judge is then restricted to assessing visual quality and semantic alignment with the task instruction. To reduce judge bias, no model evaluates its own outputs: Gemini-2.5-Pro judges GPT-4o outputs, while GPT-4o judges outputs from other models. We include the judge rubric in Tab. \ref{tab:visual-eval-structure}.

We further ensure reproducibility by fixing and logging the core execution parameters for every run, including the action space, observation type, screen resolution, execution delay, task timeout, maximum step budget, maximum trajectory length, and model decoding settings. In our setup, agents use a PyAutoGUI action space, a 2.0-second post-action delay, a 1800-second task timeout, temperature of 0.1, and top-p of 0.9. Each task is defined by a complete JSON configuration specifying the instruction, required environment setup, datasets, tools, libraries, and a custom deterministic evaluation function, making each task independently executable and verifiable. To assess run-to-run stability, we repeated the Screenshot+A11y Tree evaluation for GPT-4o and Claude-4.5-Sonnet; the overall accuracies remained within ±1\%, with GPT-4o achieving 24.54\% and 23.63\%, and Claude-4.5-Sonnet achieving 9.21\% and 9.81\%. This indicates that the benchmark pipeline produces stable and reproducible measurements.
\begin{table}[h]
\centering
\caption{Structure of a deterministic numerical evaluation function.}
\label{tab:numerical-eval-structure}

\resizebox{\columnwidth}{!}{%
\begin{tabular}{l|p{9cm}}
\toprule
\rowcolor{gray!30}
\textbf{Evaluation Step} & \textbf{Description} \\
\midrule
\rowcolor{gray!5}
Script Validation 
& Confirms that the required Python script exists and executes successfully. \\
\rowcolor{gray!5}
Output Verification
& Verifies that the expected output file (e.g., \texttt{.txt} or \texttt{.csv}) is generated. \\
\rowcolor{gray!5}
Value Extraction 
& Extracts numeric values using robust parsing to avoid formatting artifacts. \\
\rowcolor{gray!5}
Numerical Matching 
& Compares extracted values against ground-truth references using a small tolerance to account for floating-point variation. \\
\rowcolor{gray!5}
Scoring 
& Assigns partial credit for intermediate correctness and caps the final score at 1.0. \\
\bottomrule
\end{tabular}
}
\end{table}
\begin{table*}[t]
\centering
\caption{\DSAgentBench{} \textbf{average score (\%) }across ds lifecycle task types and overall performance under Screenshot and Screenshot + Accessibility Tree observation settings.}
\resizebox{\textwidth}{!}{%
\begin{tabular}{l|ccccccc|ccccccc}
\toprule
\multirow{2}{*}{\textbf{Model}}
& \multicolumn{7}{c|}{\textbf{Screenshot}}
& \multicolumn{7}{c}{\textbf{Screenshot + A11y Tree}} \\

\cmidrule(lr){2-8} \cmidrule(lr){9-15}

& \textbf{DA}
& \textbf{EDA}
& \textbf{FE}
& \textbf{Model}
& \textbf{Vis}
& \textbf{Eval}
& \textbf{Overall}

& \textbf{DA}
& \textbf{EDA}
& \textbf{FE}
& \textbf{Model}
& \textbf{Vis}
& \textbf{Eval}
& \textbf{Overall} \\

\midrule
\multicolumn{15}{l}{\textbf{\textit{Closed-Source Models}}} \\

\rowcolor{closed_models!50}
GPT4-o
& 0.26 & \textbf{0.59} & \textbf{0.63} & \textbf{0.50} & 0.48 & \textbf{0.55} & \textbf{0.53}
& \textbf{0.37} & \textbf{0.58} & \textbf{0.61} & \textbf{0.52} & 0.60 & \textbf{0.39} & \textbf{0.55} \\

\rowcolor{closed_models!50}
O4-mini
& 0.28 & 0.27 & 0.35 & 0.31 & 0.37 & 0.16 & 0.30
& 0.26 & 0.37 & 0.42 & 0.37 & 0.42 & 0.31 & 0.37 \\

\rowcolor{closed_models!50}
GPT5-mini
& \textbf{0.31} & 0.49 & 0.55 & 0.45 & \textbf{0.55} & 0.32 & 0.47
& 0.27 & 0.49 & 0.57 & 0.44 & 0.58 & 0.31 & 0.48 \\

\rowcolor{closed_models!50}
GPT5
&0.55  & 0.60 &0.62  & 0.61 & 0.58 & 0.47 & 0.59
& 0.65 & 0.67 & 0.63 & 0.66 & 0.67 & 0.40 & 0.65 \\

\rowcolor{closed_models!50}
Gemini-Pro-2.5
& 0.27 & 0.48 & .47 & 0.48 & 0.46 & 0.48 & 0.45
& 0.25 & 0.51 & 0.61 & 0.45 & \textbf{0.68} & 0.38 & 0.50 \\

\rowcolor{closed_models!50}
OpenAI CUA
& 0.22 & 0.43 & 0.46 & 0.41 & 0.51 & 0.32 & 0.42
& 0.27 & 0.40 & 0.37 & 0.40 & 0.44 & 0.34 & 0.39 \\

\rowcolor{closed_models!50}
Claude-4-Sonnet
& 0.22 & 0.37 & 0.37 & 0.39 & 0.37 & 0.27 & 0.35
& 0.24 & 0.37 & 0.35 & 0.39 & 0.45 & 0.17 & 0.36 \\

\rowcolor{closed_models!50}
Claude-4.5-Sonnet
& 0.26 & 0.36 & 0.36 & 0.39 & 0.47 & 0.36 & 0.37
& 0.28 & 0.41 & 0.46 & 0.37 & 0.46 & 0.50 & 0.41 \\

\rowcolor{closed_models!50}
Claude-4.6-Sonnet
& 0.68 & 0.72 & 0.70 & 0.74 & 0.75 & 0.71 & 0.72
& 0.64 &0.80  & 0.77 &0.76  & 0.75 & 0.72 & 0.76 \\

\midrule
\multicolumn{15}{l}{\textbf{\textit{Open-Source Models}}} \\

\rowcolor{open_models_below_4B!50}
Jedi-3B w/GPT4o
& 0.22 & 0.31 & 0.30 & 0.34 & 0.35 & 0.31 & 0.31
& N/A & N/A & N/A & N/A & N/A & N/A & N/A \\

\rowcolor{open_models_below_4B!50}
Jedi-7B w/GPT4o
& 0.20 & 0.30 & 0.31 & 0.34 & 0.37 & 0.31 & 0.31
& N/A & N/A & N/A & N/A & N/A & N/A & N/A \\

\rowcolor{open_models_7B_12B!50}
UI-Tars-2B
& 0.08 & 0.06 & 0.03 & 0.08 & 0.12 & 0.10 & 0.07
& N/A & N/A & N/A & N/A & N/A & N/A & N/A \\

\rowcolor{open_models_7B_12B!50}
UI-Tars-1.5-7B
& 0.00 & 0.02 & 0.00 & 0.03 & 0.03 & 0.02 & 0.02
& N/A & N/A & N/A & N/A & N/A & N/A & N/A \\

\rowcolor{open_models_7B_12B!50}
GUI-OWL-7B
& 0.03 & 0.05 & 0.08 & 0.04 & 0.01 & 0.08 & 0.05
& N/A & N/A & N/A & N/A & N/A & N/A & N/A \\

\rowcolor{open_models_7B_12B!50}
OpenCUA 72b
& 0.10 & 0.09 & 0.09 & 0.09 & 0.18 & 0.06 & 0.10
& N/A & N/A & N/A & N/A & N/A & N/A & N/A \\

\bottomrule
\end{tabular}
}
\label{tab:dsagentbench-lifecycle_avg}
\end{table*}

\section{ Evaluation Functions}
\label{app:evaluation}

\subsection{ Numerical Result Verification}

This evaluator targets tasks with a known ground-truth numerical answer (e.g., correlation, volatility, or aggregate counts). It verifies correct script execution and numerical accuracy within a predefined tolerance (see Tab. \ref{tab:numerical-eval-structure}).

This evaluator is fully deterministic and does not rely on language models, ensuring stable and repeatable scores across runs.

\begin{table}[h]
\centering
\caption{Visualization quality evaluation description.}

\label{tab:visual-eval-structure}

\resizebox{\columnwidth}{!}{%
\begin{tabular}{l|p{9cm}}
\toprule
\rowcolor{gray!30}
\textbf{Evaluation Step} & \textbf{Description} \\
\midrule
\rowcolor{gray!5}
Script Validation 
& Ensures that the visualization script exists and executes successfully. \\
\rowcolor{gray!5}
Artifact Verification 
& Confirms that required visualization files are generated and non-empty. \\
\rowcolor{gray!5}
Semantic Validation 
& Uses a fixed evaluation prompt with GPT-4o as a judge to assess whether each chart correctly represents the intended variables, axes, and trends. \\
\rowcolor{gray!5}
Design Criteria 
& Checks readability, labeling, scale consistency, and layout clarity. \\
\rowcolor{gray!5}
Scoring 
& Combines deterministic and semantic signals into a bounded score in the range [0, 1]. \\
\bottomrule
\end{tabular}
}
\end{table}

\subsection{Visualization Quality Evaluation}

For visualization tasks, correctness cannot be assessed purely numerically. We therefore combine deterministic checks with grounded semantic validation to assess chart clarity and representational fidelity (see Tab. \ref{tab:visual-eval-structure}). This evaluator captures semantic correctness while minimizing subjectivity through constrained prompts and binary judgments.



\section{Additional Results}
We provide more result analysis in this section (see Table~\ref{tab:trajectory-analysis}, Table~\ref{tab:dsagentbench-open}, and Table~\ref{tab:dsagentbench-lifecycle_avg}).

\subsection{Error Analysis Examples}
\begin{figure*}[t]
    \includegraphics[width=.5\textwidth]{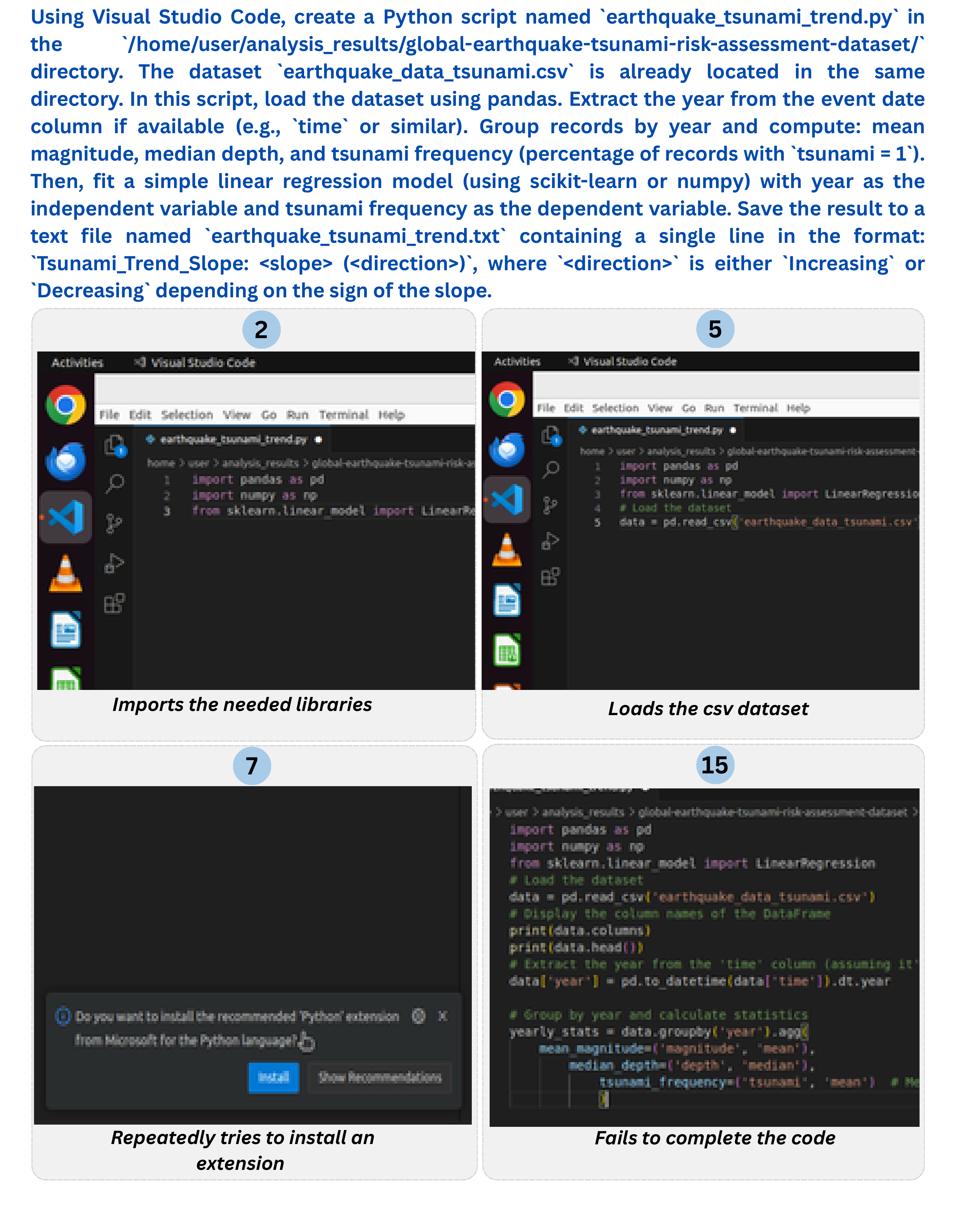}
    \caption{Example of error that occured with the model OpenAI CUA.   
    }
    \label{fig:error_cua}
\end{figure*}

\begin{figure*}[t]
    \includegraphics[width=.5\textwidth]{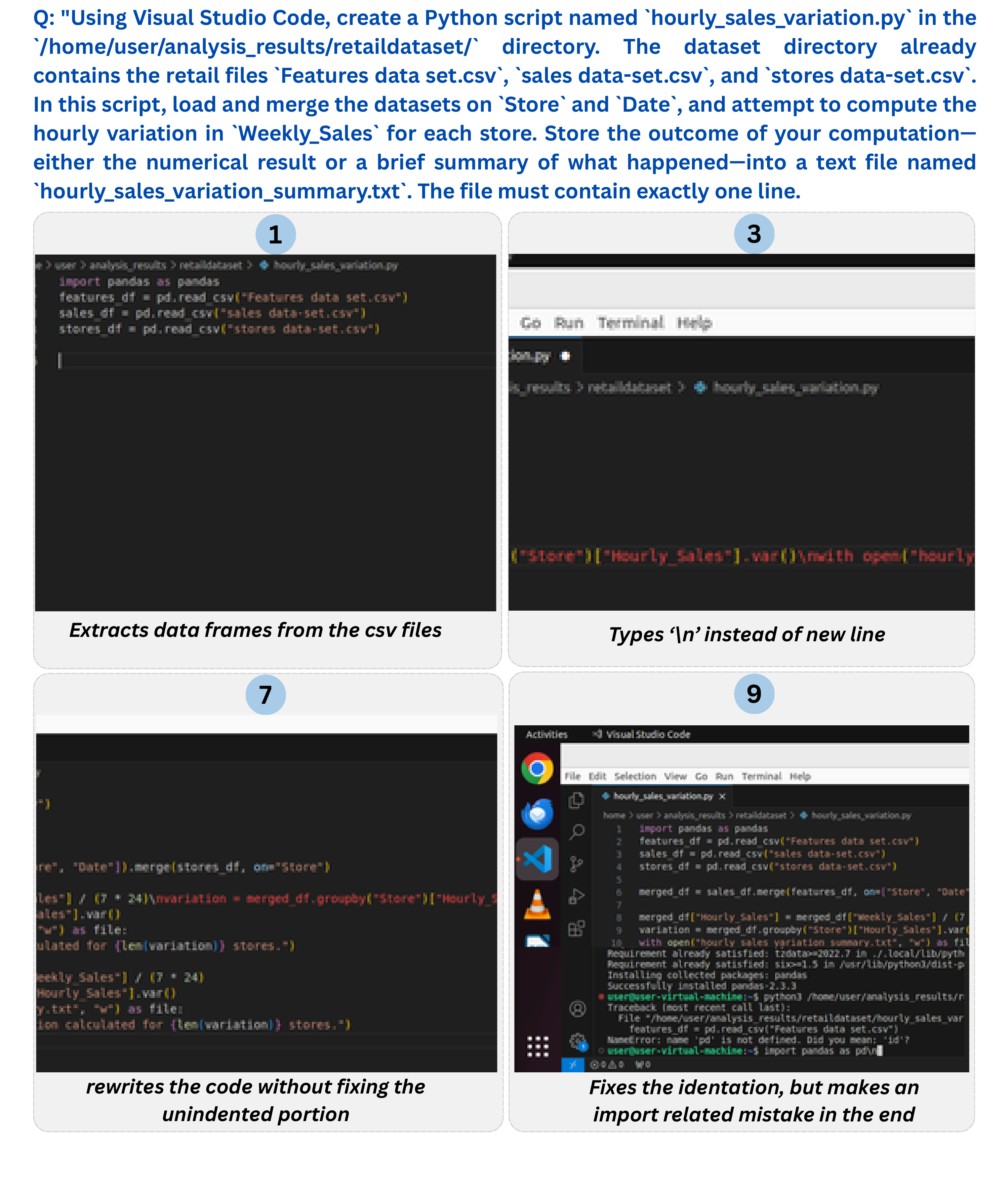}
    \caption{Example of error that occured with the model Jedi-3B.   
    }
    \label{fig:error_j3}
\end{figure*}

\begin{figure*}[t]
    \includegraphics[width=.5\textwidth]{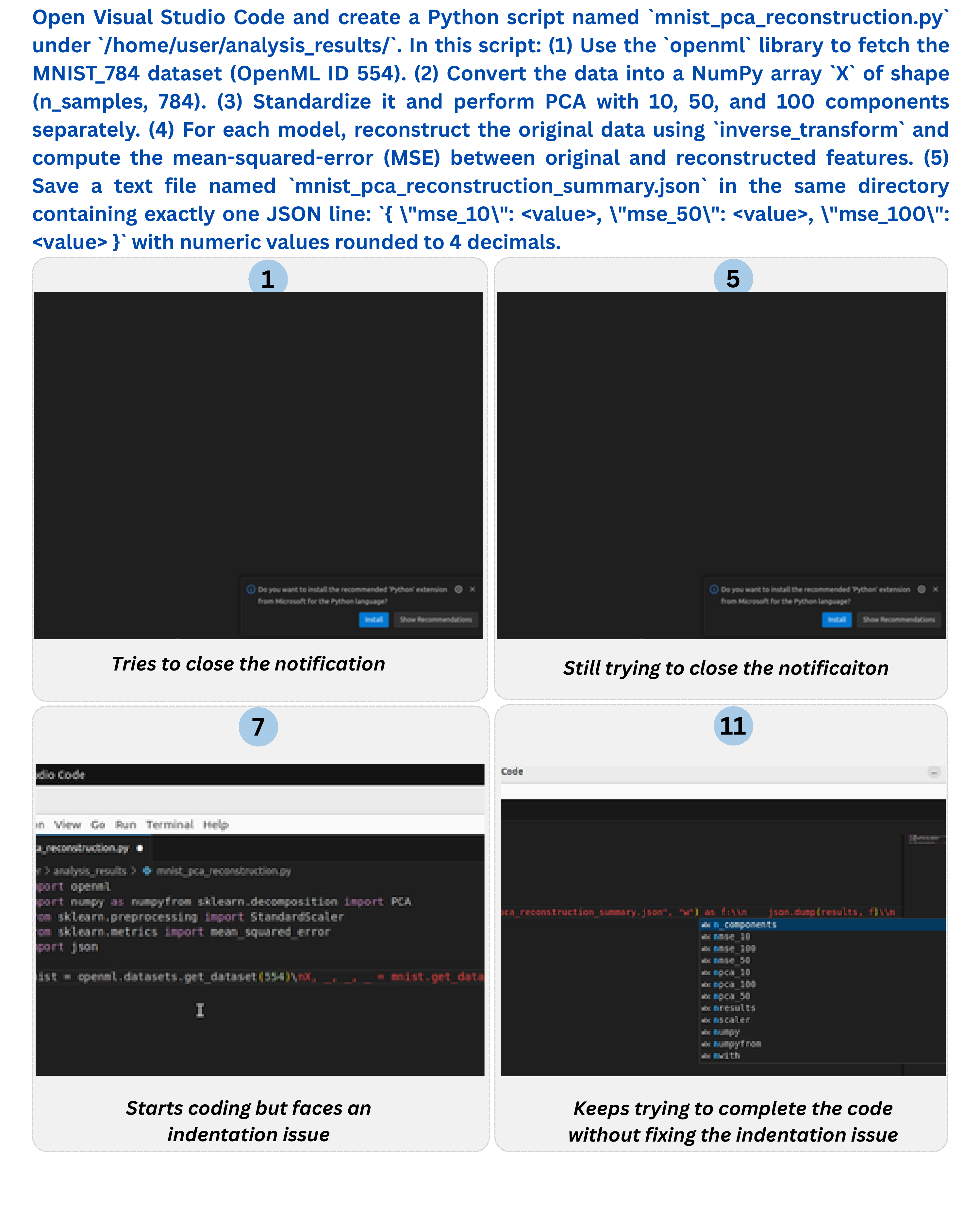}
    \caption{Example of error that occured with the model Jedi-8B.   
    }
    \label{fig:error_j8}
\end{figure*}

\begin{figure*}[t]
    \includegraphics[width=.5\textwidth]{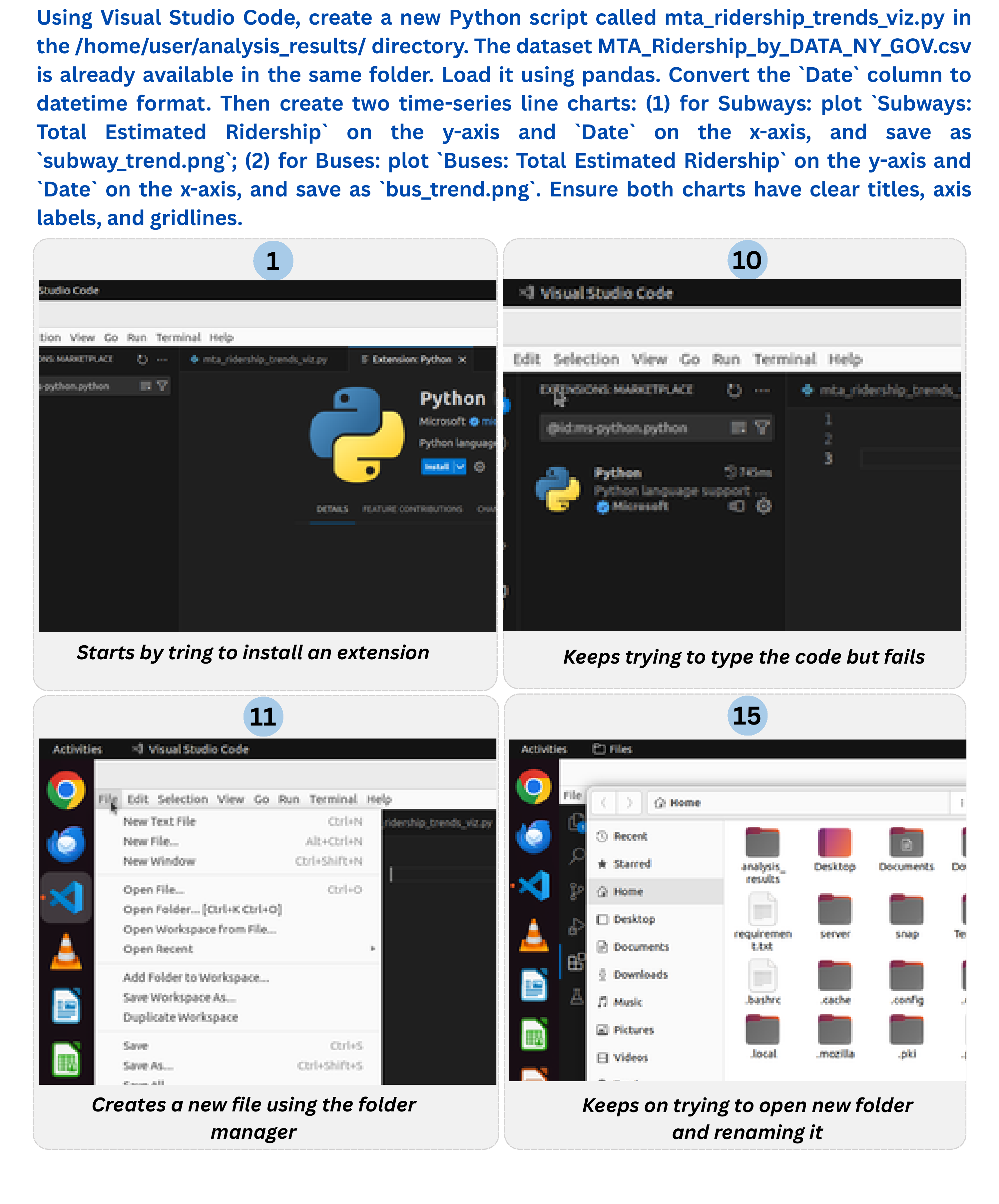}
    \caption{Example of error that occured with the model UI-Tars-1.5-7B.   
    }
    \label{fig:error_tars}
\end{figure*}

Additional qualitative error cases produced by different models are illustrated in \cref{fig:error_cua}, \cref{fig:error_j3}, \cref{fig:error_j8}, and \cref{fig:error_tars}. Across these examples, we observe several recurring failure patterns that are consistent across model families.

A prominent source of errors arises from the models’ difficulty in handling environmental notifications and system-level UI elements. In particular, the models frequently fail to correctly interpret or dismiss generic pop-up notifications originating from the code editor environment, which subsequently disrupts task progression and leads to incorrect or incomplete actions.

In addition, the models exhibit systematic weaknesses in maintaining proper code formatting and indentation. We observe multiple instances where a model intends to insert a line break but instead explicitly generates the string token \textbackslash n within the code. Since this token is not interpreted as an actual newline by the execution environment, it results in syntactically invalid or semantically incorrect code. This mismatch between the model’s internal representation of formatting and the execution semantics ultimately propagates downstream errors and task failure.

Overall, these error patterns highlight limitations in current models’ robustness to interactive development environments and their ability to reliably translate formatting intentions into executable code.

\begin{figure*}[t]
    \centering
    
    \begin{tcolorbox}[
        colback=gray!5!white,
        colframe=gray!60!black,
        title=\textbf{Standard System Prompt (Screenshot Input $\to$ Code Output)},
        fonttitle=\bfseries\small,
        boxrule=0.8pt,
        arc=2pt,
        left=2pt, right=2pt, top=2pt, bottom=2pt
    ]
    \fontfamily{cmtt}\selectfont\small
    You are an agent which follow my instruction and perform desktop computer tasks as instructed. \\
    You have good knowledge of computer and good internet connection and assume your code will run on a computer for controlling the mouse and keyboard. \\
    For each step, you will get an observation of an image, which is the screenshot of the computer screen and you will predict the action of the computer based on the image. \\
    
    You are required to use `pyautogui` to perform the action grounded to the observation, but DONOT use the `pyautogui.locateCenterOnScreen` function to locate the element you want to operate with since we have no image of the element you want to operate with. DONOT USE `pyautogui.screenshot()` to make screenshot. \\
    Return one line or multiple lines of python code to perform the action each time, be time efficient. When predicting multiple lines of code, make some small sleep like `time.sleep(0.5);` interval so that the machine could take; Each time you need to predict a complete code, no variables or function can be shared from history \\
    You need to to specify the coordinates of by yourself based on your observation of current observation, but you should be careful to ensure that the coordinates are correct. \\
    You ONLY need to return the code inside a code block, like this: \\
    ```python \\
    \# your code here \\
    ``` \\
    Specially, it is also allowed to return the following special code: \\
    When you think you have to wait for some time, return ```WAIT```; \\
    When you think the task can not be done, return ```FAIL```, don't easily say ```FAIL```, try your best to do the task; \\
    When you think the task is done, return ```DONE```. \\
    
    My computer's password is '\{CLIENT\_PASSWORD\}', feel free to use it when you need sudo rights. \\
    First give the current screenshot and previous things we did a short reflection, then RETURN ME THE CODE OR SPECIAL CODE I ASKED FOR. NEVER EVER RETURN ME ANYTHING ELSE.
    \end{tcolorbox}
    
    \vspace{0.2cm}
    
    \begin{tcolorbox}[
        colback=gray!5!white,
        colframe=gray!60!black,
        title=\textbf{Few-Shot System Prompt (Screenshot Input $\to$ Code Output)},
        fonttitle=\bfseries\small,
        boxrule=0.8pt,
        arc=2pt,
        left=2pt, right=2pt, top=2pt, bottom=2pt
    ]
    \fontfamily{cmtt}\selectfont\small
    You are an agent which follow my instruction and perform desktop computer tasks as instructed. \\
    You have good knowledge of computer and good internet connection and assume your code will run on a computer for controlling the mouse and keyboard. \\
    For each step, you will get an observation of an image, which is the screenshot of the computer screen and the instruction and you will predict the next action to operate on the computer based on the image. \\
    
    You are required to use `pyautogui` to perform the action grounded to the observation, but DONOT use the `pyautogui.locateCenterOnScreen` function to locate the element you want to operate with since we have no image of the element you want to operate with. DONOT USE `pyautogui.screenshot()` to make screenshot. \\
    Return one line or multiple lines of python code to perform the action each time, be time efficient. When predicting multiple lines of code, make some small sleep like `time.sleep(0.5);` interval so that the machine could take; Each time you need to predict a complete code, no variables or function can be shared from history \\
    You need to to specify the coordinates of by yourself based on your observation of current observation, but you should be careful to ensure that the coordinates are correct. \\
    You ONLY need to return the code inside a code block, like this: \\
    ```python \\
    \# your code here \\
    ``` \\
    Specially, it is also allowed to return the following special code: \\
    When you think you have to wait for some time, return ```WAIT```; \\
    When you think the task can not be done, return ```FAIL```, don't easily say ```FAIL```, try your best to do the task; \\
    When you think the task is done, return ```DONE```. \\
    
    My computer's password is '\{CLIENT\_PASSWORD\}', feel free to use it when you need sudo rights. \\
    Our past communication is great, and what you have done is very helpful. I will now give you another task to complete. \\
    First take a deep breath, think step by step, give the current screenshot a thinking, then RETURN ME THE CODE OR SPECIAL CODE I ASKED FOR. NEVER EVER RETURN ME ANYTHING ELSE.
    \end{tcolorbox}

    \caption{The system prompts used for the code-generation agent. The \textbf{Standard Prompt} (top) focuses on direct execution, while the \textbf{Few-Shot Prompt} (bottom) incorporates "deep breath" CoT triggering.}
    \label{fig:screenshot_code_prompts}
\end{figure*}

\begin{figure*}[t]
    \centering
    
    \begin{tcolorbox}[
        colback=gray!5!white, 
        colframe=gray!60!black, 
        title=\textbf{UITARS User Prompt (No Thought)},
        fonttitle=\bfseries\small,
        boxrule=0.8pt,
        arc=2pt
    ]
    \fontfamily{cmtt}\selectfont\footnotesize
    You are a GUI agent. You are given a task and your action history, with screenshots. You need to perform the next action to complete the task. \\
    \textbf{\#\# Output Format} \\
    ``` \\
    Action: ... \\
    ``` \\
    \textbf{\#\# Action Space} \\
    click(start\_box='<|box\_start|>(x1,y1)<|box\_end|>') \\
    left\_double(start\_box='<|box\_start|>(x1,y1)<|box\_end|>') \\
    right\_single(start\_box='<|box\_start|>(x1,y1)<|box\_end|>') \\
    drag(start\_box='<|box\_start|>(x1,y1)<|box\_end|>', end\_box='<|box\_start|>(x3,y3)<|box\_end|>') \\
    hotkey(key='') \\
    type(content='') \#If you want to submit your input, use "$\backslash$n" at the end of `content`. \\
    scroll(start\_box='<|box\_start|>(x1,y1)<|box\_end|>', direction='down or up or right or left') \\
    wait() \#Sleep for 5s and take a screenshot to check for any changes. \\
    finished() \\
    call\_user() \# Submit the task and call the user when the task is unsolvable, or when you need the user's help. \\
    
    \textbf{\#\# User Instruction} \\
    \{instruction\}
    \end{tcolorbox}
    
    \vspace{0.3cm} 

    \begin{tcolorbox}[
        colback=gray!5!white,
        colframe=gray!60!black,
        title=\textbf{UITARS User Prompt (Thought)},
        fonttitle=\bfseries\small,
        boxrule=0.8pt,
        arc=2pt
    ]
    \fontfamily{cmtt}\selectfont\footnotesize
    You are a GUI agent. You are given a task and your action history, with screenshots. You need to perform the next action to complete the task. \\
    
    \textbf{\#\# Output Format} \\
    ``` \\
    Thought: ... \\
    Action: ... \\
    ``` \\
    
    \textbf{\#\# Action Space} \\
    \{action\_space\} \\
    
    \textbf{\#\# Note} \\
    - Use \{language\} in `Thought` part. \\
    - Write a small plan and finally summarize your next action (with its target element) in one sentence in `Thought` part. \\
    
    \textbf{\#\# User Instruction} \\
    \{instruction\}
    \end{tcolorbox}

    \caption{The system prompts used for the UITARS agent. The top box shows the standard prompting strategy, while the bottom box includes `Thought' reasoning requirements.}
    \label{fig:uitars_prompts}
\end{figure*}

\begin{figure*}[ht] 
    \centering
    
    \begin{tcolorbox}[
        colback=gray!5!white,
        colframe=gray!60!black,
        title=\textbf{JEDI Grounder System Prompt},
        fonttitle=\bfseries\small,
        boxrule=0.8pt,
        arc=2pt,
        left=2pt, right=2pt, top=2pt, bottom=2pt
    ]
    \fontfamily{cmtt}\selectfont\small
    You are a helpful assistant. \\
    
    \textbf{\# Tools} \\
    
    You may call one or more functions to assist with the user query. \\
    
    You are provided with function signatures within \textless tools\textgreater\ XML tags: \\
    \textless tools\textgreater \\
    \{\{"type": "function", "function": \{\{"name": "computer\_use", "description": "Use a mouse and keyboard to interact with a computer, and take screenshots.\textbackslash n* This is an interface to a desktop GUI. You do not have access to a terminal or applications menu. You must click on desktop icons to start applications.\textbackslash n* Some applications may take time to start or process actions, so you may need to wait and take successive screenshots to see the results of your actions. E.g. if you click on Firefox and a window doesn't open, try wait and taking another screenshot.\textbackslash n* The screen's resolution is \{width\}x\{height\}.\textbackslash n* Whenever you intend to move the cursor to click on an element like an icon, you should consult a screenshot to determine the coordinates of the element before moving the cursor.\textbackslash n* If you tried clicking on a program or link but it failed to load, even after waiting, try adjusting your cursor position so that the tip of the cursor visually falls on the element that you want to click.\textbackslash n* Make sure to click any buttons, links, icons, etc with the cursor tip in the center of the element. Don't click boxes on their edges unless asked.", "parameters": \{\{"properties": \{\{"action": \{\{"description": "The action to perform. The available actions are:\textbackslash n* `key`: Performs key down presses on the arguments passed in order, then performs key releases in reverse order.\textbackslash n* `type`: Type a string of text on the keyboard.\textbackslash n* `mouse\_move`: Move the cursor to a specified (x, y) pixel coordinate on the screen.\textbackslash n* `left\_click`: Click the left mouse button.\textbackslash n* `left\_click\_drag`: Click and drag the cursor to a specified (x, y) pixel coordinate on the screen.\textbackslash n* `right\_click`: Click the right mouse button.\textbackslash n* `middle\_click`: Click the middle mouse button.\textbackslash n* `double\_click`: Double-click the left mouse button.\textbackslash n* `scroll`: Performs a scroll of the mouse scroll wheel.\textbackslash n* `wait`: Wait specified seconds for the change to happen.\textbackslash n* `terminate`: Terminate the current task and report its completion status.", "enum": {[}"key", "type", "mouse\_move", "left\_click", "left\_click\_drag", "right\_click", "middle\_click", "double\_click", "scroll", "wait", "terminate"{]}, "type": "string"\}\}, "keys": \{\{"description": "Required only by `action=key`.", "type": "array"\}\}, "text": \{\{"description": "Required only by `action=type`.", "type": "string"\}\}, "coordinate": \{\{"description": "(x, y): The x (pixels from the left edge) and y (pixels from the top edge) coordinates to move the mouse to. Required only by `action=mouse\_move`, `action=left\_click\_drag`, `action=left\_click`, `action=right\_click`, `action=double\_click`.", "type": "array"\}\}, "pixels": \{\{"description": "The amount of scrolling to perform. Positive values scroll up, negative values scroll down. Required only by `action=scroll`.", "type": "number"\}\}, "time": \{\{"description": "The seconds to wait. Required only by `action=wait`.", "type": "number"\}\}, "status": \{\{"description": "The status of the task. Required only by `action=terminate`.", "type": "string", "enum": {[}"success", "failure"{]} \}\}\}\}, "required": {[}"action"{]}, "type": "object"\}\}\}\}\}\} \\
    \textless /tools\textgreater \\
    
    For each function call, return a json object with function name and arguments within \textless tool\_call\textgreater\ XML tags: \\
    \textless tool\_call\textgreater \\
    \{\{"name": \textless function-name\textgreater, "arguments": \textless args-json-object\textgreater\}\} \\
    \textless /tool\_call\textgreater
    \end{tcolorbox}

    \caption{Example system prompt for the JEDI architecture. The Grounder prompt provides the precise tool definitions and XML schema for tool calling.}
    \label{fig:jedi_ground_prompt}
\end{figure*}

\begin{figure*}[t]
    \centering
    
    \begin{tcolorbox}[
        colback=gray!5!white,
        colframe=gray!60!black,
        title=\textbf{JEDI Planner System Prompt},
        fonttitle=\bfseries\small,
        boxrule=0.8pt,
        arc=2pt,
        left=2pt, right=2pt, top=2pt, bottom=2pt
    ]
    \fontfamily{cmtt}\selectfont\scriptsize
    You are an agent which follow my instruction and perform desktop computer tasks as instructed. \\
    You have good knowledge of computer and good internet connection and assume your code will run on a computer for controlling the mouse and keyboard. \\
    
    You are on Ubuntu operating system and the resolution of the screen is 1920x1080. \\
    For each step, you will get an observation of an image, which is the screenshot of the computer screen and you will predict the action of the computer based on the image. \\
    
    The following rules are IMPORTANT: \\
    - If previous actions didn't achieve the expected result, do not repeat them, especially the last one. Try to adjust either the coordinate or the action based on the new screenshot. \\
    - Do not predict multiple clicks at once. Base each action on the current screenshot; do not predict actions for elements or events not yet visible in the screenshot. \\
    - You cannot complete the task by outputting text content in your response. You must use mouse and keyboard to interact with the computer. Return ```Fail``` when you think the task can not be done. \\
    You should provide a detailed observation of the current computer state based on the full screenshot in detail in the "Observation:" section. \\
    Provide any information that is possibly relevant to achieving the task goal and any elements that may affect the task execution, such as pop-ups, notifications, error messages, loading states, etc.. \\
    You MUST return the observation before the thought. \\
    You should think step by step and provide a detailed thought process before generating the next action: \\
    Thought: \\
    - Step by Step Progress Assessment: \\
    \hspace*{0.5cm} - Analyze completed task parts and their contribution to the overall goal \\
    \hspace*{0.5cm} - Reflect on potential errors, unexpected results, or obstacles \\
    \hspace*{0.5cm} - If previous action was incorrect, predict a logical recovery step \\
    - Next Action Analysis: \\
    \hspace*{0.5cm} - List possible next actions based on current state \\
    \hspace*{0.5cm} - Evaluate options considering current state and previous actions \\
    \hspace*{0.5cm} - Propose most logical next action \\
    \hspace*{0.5cm} - Anticipate consequences of the proposed action \\
    
    Your thought should be returned in "Thought:" section. You MUST return the thought before the code. \\
    
    You are required to use `pyautogui` to perform the action grounded to the observation, but DONOT use the `pyautogui.locateCenterOnScreen` function to locate the element you want to operate with since we have no image of the element you want to operate with. DONOT USE `pyautogui.screenshot()` to make screenshot. \\
    Return exactly ONE line of python code to perform the action each time. At each step, you MUST generate the corresponding instruction to the code before a \# in a comment (example: \# Click \"Yes, I trust the authors\" button\textbackslash npyautogui.click(x=0, y=0, duration=1)\textbackslash n) \\
    
    For the instruction you can decribe the element you want to interact with in detail including the visual description and function description. And make it clear and concise. \\
    For example you can describe what the element looks like, and what will be the expected result when you interact with it. \\
    You need to to specify the coordinates of by yourself based on your observation of current observation, but you should be careful to ensure that the coordinates are correct. \\
    Remember you should only return ONE line of code, DO NOT RETURN more. You should return the code inside a code block, like this: \\
    ```python \\
    \# your code here \\
    ``` \\
    Specially, it is also allowed to return the following special code: \\
    When you think you have to wait for some time, return ```WAIT```; \\
    When you think the task can not be done, return ```FAIL```, don't easily say ```FAIL```, try your best to do the task; \\
    When you think the task is done, return ```DONE```. \\
    
    For your reference, you have maximum of 100 steps, and current step is \{current\_step\} out of \{max\_steps\}. \\
    If you are in the last step, you should return ```DONE``` or ```FAIL``` according to the result. \\
    
    Here are some guidelines for you: \\
    1. Remember to generate the corresponding instruction to the code before a \# in a comment and only return ONE line of code. \\
    2. If a click action is needed, use only the following functions: pyautogui.click, pyautogui.rightClick or pyautogui.doubleClick. \\
    3. Return ```Done``` when you think the task is done. Return ```Fail``` when you think the task can not be done. \\
    
    My computer's password is '\{CLIENT\_PASSWORD\}', feel free to use it when you need sudo rights. \\
    First give the current screenshot and previous things we did a short reflection, then RETURN ME THE CODE OR SPECIAL CODE I ASKED FOR NEVER EVER RETURN ME ANYTHING ELSE.
    \end{tcolorbox}
    \caption{Example system prompt for the JEDI architecture. The Planner prompt enforces a strict structure for Observation, `Thought` reasoning, and Python code generation.}
    \label{fig:jedi_planner_prompt}
\end{figure*}

%% file: emnlp2020-templates/dashInteractQA.bib
@article{zhang2020data,
  title={How do data science workers collaborate? roles, workflows, and tools},
  author={Zhang, Amy X and Muller, Michael and Wang, Dakuo},
  journal={Proceedings of the ACM on Human-Computer Interaction},
  volume={4},
  number={CSCW1},
  pages={1--23},
  year={2020},
  publisher={ACM New York, NY, USA}
}

@article{li2024autokaggle,
  title={Autokaggle: A multi-agent framework for autonomous data science competitions},
  author={Li, Ziming and Zang, Qianbo and Ma, David and Guo, Jiawei and Zheng, Tuney and Liu, Minghao and Niu, Xinyao and Wang, Yue and Yang, Jian and Liu, Jiaheng and others},
  journal={arXiv preprint arXiv:2410.20424},
  year={2024}
}

@article{qiao2023taskweaver,
  title={Taskweaver: A code-first agent framework},
  author={Qiao, Bo and Li, Liqun and Zhang, Xu and He, Shilin and Kang, Yu and Zhang, Chaoyun and Yang, Fangkai and Dong, Hang and Zhang, Jue and Wang, Lu and others},
  journal={arXiv preprint arXiv:2311.17541},
  year={2023}
}

@inproceedings{Guo-2024-DS-agent,
author = {Guo, Siyuan and Deng, Cheng and Wen, Ying and Chen, Hechang and Chang, Yi and Wang, Jun},
title = {DS-agent: automated data science by empowering large language models with case-based reasoning},
year = {2024},
publisher = {JMLR.org},
booktitle = {Proceedings of the 41st International Conference on Machine Learning},
articleno = {668},
numpages = {36},
location = {Vienna, Austria},
series = {ICML'24}
}

@article{masry2022chartqa,
  title={Chartqa: A benchmark for question answering about charts with visual and logical reasoning},
  author={Masry, Ahmed and Long, Do Xuan and Tan, Jia Qing and Joty, Shafiq and Hoque, Enamul},
  journal={arXiv preprint arXiv:2203.10244},
  year={2022}
}

@article{xie2024osworld,
  title={Osworld: Benchmarking multimodal agents for open-ended tasks in real computer environments},
  author={Xie, Tianbao and Zhang, Danyang and Chen, Jixuan and Li, Xiaochuan and Zhao, Siheng and Cao, Ruisheng and Hua, Toh J and Cheng, Zhoujun and Shin, Dongchan and Lei, Fangyu and others},
  journal={Advances in Neural Information Processing Systems},
  volume={37},
  pages={52040--52094},
  year={2024}
}

@misc{jedi,
      title={Scaling Computer-Use Grounding via User Interface Decomposition and Synthesis}, 
      author={Tianbao Xie and Jiaqi Deng and Xiaochuan Li and Junlin Yang and Haoyuan Wu and Jixuan Chen and Wenjing Hu and Xinyuan Wang and Yuhui Xu and Zekun Wang and Yiheng Xu and Junli Wang and Doyen Sahoo and Tao Yu and Caiming Xiong},
      year={2025},
      eprint={2505.13227},
      archivePrefix={arXiv},
      primaryClass={cs.AI},
      url={https://arxiv.org/abs/2505.13227}, 
}

@misc{uitars,
      title={UI-TARS: Pioneering Automated GUI Interaction with Native Agents}, 
      author={Yujia Qin and Yining Ye and Junjie Fang and Haoming Wang and Shihao Liang and Shizuo Tian and Junda Zhang and Jiahao Li and Yunxin Li and Shijue Huang and Wanjun Zhong and Kuanye Li and Jiale Yang and Yu Miao and Woyu Lin and Longxiang Liu and Xu Jiang and Qianli Ma and Jingyu Li and Xiaojun Xiao and Kai Cai and Chuang Li and Yaowei Zheng and Chaolin Jin and Chen Li and Xiao Zhou and Minchao Wang and Haoli Chen and Zhaojian Li and Haihua Yang and Haifeng Liu and Feng Lin and Tao Peng and Xin Liu and Guang Shi},
      year={2025},
      eprint={2501.12326},
      archivePrefix={arXiv},
      primaryClass={cs.AI},
      url={https://arxiv.org/abs/2501.12326}, 
}

@misc{openaiComputerUsingAgent,
	author = {},
	title = {{C}omputer-{U}sing {A}gent --- openai.com},
	howpublished = {\url{https://openai.com/index/computer-using-agent/}},
	year = {},
	note = {[Accessed 14-07-2025]},
}

@misc{openai2024gpt4technicalreport,
      title={GPT-4 Technical Report}, 
      author={OpenAI and Josh Achiam and Steven Adler and Sandhini Agarwal and Lama Ahmad and Ilge Akkaya and Florencia Leoni Aleman and Diogo Almeida and Janko Altenschmidt and Sam Altman and Shyamal Anadkat and Red Avila and Igor Babuschkin and Suchir Balaji and Valerie Balcom and Paul Baltescu and Haiming Bao and Mohammad Bavarian and Jeff Belgum and Irwan Bello and Jake Berdine and Gabriel Bernadett-Shapiro and Christopher Berner and Lenny Bogdonoff and Oleg Boiko and Madelaine Boyd and Anna-Luisa Brakman and Greg Brockman and Tim Brooks and Miles Brundage and Kevin Button and Trevor Cai and Rosie Campbell and Andrew Cann and Brittany Carey and Chelsea Carlson and Rory Carmichael and Brooke Chan and Che Chang and Fotis Chantzis and Derek Chen and Sully Chen and Ruby Chen and Jason Chen and Mark Chen and Ben Chess and Chester Cho and Casey Chu and Hyung Won Chung and Dave Cummings and Jeremiah Currier and Yunxing Dai and Cory Decareaux and Thomas Degry and Noah Deutsch and Damien Deville and Arka Dhar and David Dohan and Steve Dowling and Sheila Dunning and Adrien Ecoffet and Atty Eleti and Tyna Eloundou and David Farhi and Liam Fedus and Niko Felix and Simón Posada Fishman and Juston Forte and Isabella Fulford and Leo Gao and Elie Georges and Christian Gibson and Vik Goel and Tarun Gogineni and Gabriel Goh and Rapha Gontijo-Lopes and Jonathan Gordon and Morgan Grafstein and Scott Gray and Ryan Greene and Joshua Gross and Shixiang Shane Gu and Yufei Guo and Chris Hallacy and Jesse Han and Jeff Harris and Yuchen He and Mike Heaton and Johannes Heidecke and Chris Hesse and Alan Hickey and Wade Hickey and Peter Hoeschele and Brandon Houghton and Kenny Hsu and Shengli Hu and Xin Hu and Joost Huizinga and Shantanu Jain and Shawn Jain and Joanne Jang and Angela Jiang and Roger Jiang and Haozhun Jin and Denny Jin and Shino Jomoto and Billie Jonn and Heewoo Jun and Tomer Kaftan and Łukasz Kaiser and Ali Kamali and Ingmar Kanitscheider and Nitish Shirish Keskar and Tabarak Khan and Logan Kilpatrick and Jong Wook Kim and Christina Kim and Yongjik Kim and Jan Hendrik Kirchner and Jamie Kiros and Matt Knight and Daniel Kokotajlo and Łukasz Kondraciuk and Andrew Kondrich and Aris Konstantinidis and Kyle Kosic and Gretchen Krueger and Vishal Kuo and Michael Lampe and Ikai Lan and Teddy Lee and Jan Leike and Jade Leung and Daniel Levy and Chak Ming Li and Rachel Lim and Molly Lin and Stephanie Lin and Mateusz Litwin and Theresa Lopez and Ryan Lowe and Patricia Lue and Anna Makanju and Kim Malfacini and Sam Manning and Todor Markov and Yaniv Markovski and Bianca Martin and Katie Mayer and Andrew Mayne and Bob McGrew and Scott Mayer McKinney and Christine McLeavey and Paul McMillan and Jake McNeil and David Medina and Aalok Mehta and Jacob Menick and Luke Metz and Andrey Mishchenko and Pamela Mishkin and Vinnie Monaco and Evan Morikawa and Daniel Mossing and Tong Mu and Mira Murati and Oleg Murk and David Mély and Ashvin Nair and Reiichiro Nakano and Rajeev Nayak and Arvind Neelakantan and Richard Ngo and Hyeonwoo Noh and Long Ouyang and Cullen O'Keefe and Jakub Pachocki and Alex Paino and Joe Palermo and Ashley Pantuliano and Giambattista Parascandolo and Joel Parish and Emy Parparita and Alex Passos and Mikhail Pavlov and Andrew Peng and Adam Perelman and Filipe de Avila Belbute Peres and Michael Petrov and Henrique Ponde de Oliveira Pinto and Michael and Pokorny and Michelle Pokrass and Vitchyr H. Pong and Tolly Powell and Alethea Power and Boris Power and Elizabeth Proehl and Raul Puri and Alec Radford and Jack Rae and Aditya Ramesh and Cameron Raymond and Francis Real and Kendra Rimbach and Carl Ross and Bob Rotsted and Henri Roussez and Nick Ryder and Mario Saltarelli and Ted Sanders and Shibani Santurkar and Girish Sastry and Heather Schmidt and David Schnurr and John Schulman and Daniel Selsam and Kyla Sheppard and Toki Sherbakov and Jessica Shieh and Sarah Shoker and Pranav Shyam and Szymon Sidor and Eric Sigler and Maddie Simens and Jordan Sitkin and Katarina Slama and Ian Sohl and Benjamin Sokolowsky and Yang Song and Natalie Staudacher and Felipe Petroski Such and Natalie Summers and Ilya Sutskever and Jie Tang and Nikolas Tezak and Madeleine B. Thompson and Phil Tillet and Amin Tootoonchian and Elizabeth Tseng and Preston Tuggle and Nick Turley and Jerry Tworek and Juan Felipe Cerón Uribe and Andrea Vallone and Arun Vijayvergiya and Chelsea Voss and Carroll Wainwright and Justin Jay Wang and Alvin Wang and Ben Wang and Jonathan Ward and Jason Wei and CJ Weinmann and Akila Welihinda and Peter Welinder and Jiayi Weng and Lilian Weng and Matt Wiethoff and Dave Willner and Clemens Winter and Samuel Wolrich and Hannah Wong and Lauren Workman and Sherwin Wu and Jeff Wu and Michael Wu and Kai Xiao and Tao Xu and Sarah Yoo and Kevin Yu and Qiming Yuan and Wojciech Zaremba and Rowan Zellers and Chong Zhang and Marvin Zhang and Shengjia Zhao and Tianhao Zheng and Juntang Zhuang and William Zhuk and Barret Zoph},
      year={2024},
      eprint={2303.08774},
      archivePrefix={arXiv},
      primaryClass={cs.CL},
      url={https://arxiv.org/abs/2303.08774}, 
}

@inproceedings{vllm,
  title={Efficient Memory Management for Large Language Model Serving with PagedAttention},
  author={Woosuk Kwon and Zhuohan Li and Siyuan Zhuang and Ying Sheng and Lianmin Zheng and Cody Hao Yu and Joseph E. Gonzalez and Hao Zhang and Ion Stoica},
  booktitle={Proceedings of the ACM SIGOPS 29th Symposium on Operating Systems Principles},
  year={2023}
}

@misc{geminipro25,
      title={Gemini 2.5: Pushing the Frontier with Advanced Reasoning, Multimodality, Long Context, and Next Generation Agentic Capabilities}, 
      author={Gheorghe Comanici and Eric Bieber and Mike Schaekermann and Ice Pasupat and Noveen Sachdeva and Inderjit Dhillon, et al.},
      year={2025},
      eprint={2507.06261},
      archivePrefix={arXiv},
      primaryClass={cs.CL},
      url={https://arxiv.org/abs/2507.06261}, 
}

@online{openai2025gpt5,
  title        = {Introducing GPT-5},
  author       = {{OpenAI}},
  year         = {2025},
  url          = {https://openai.com/index/introducing-gpt-5/},
  note         = {Accessed: 2025-10-01}
}

@misc{ye2025mobileagentv3fundamentalagentsgui,
      title={Mobile-Agent-v3: Fundamental Agents for GUI Automation}, 
      author={Jiabo Ye and Xi Zhang and Haiyang Xu and Haowei Liu and Junyang Wang and Zhaoqing Zhu and Ziwei Zheng and Feiyu Gao and Junjie Cao and Zhengxi Lu and Jitong Liao and Qi Zheng and Fei Huang and Jingren Zhou and Ming Yan},
      year={2025},
      eprint={2508.15144},
      archivePrefix={arXiv},
      primaryClass={cs.AI},
      url={https://arxiv.org/abs/2508.15144}, 
}

@article{chen2021evaluating,
  title={Evaluating large language models trained on code},
  author={Chen, Mark},
  journal={arXiv preprint arXiv:2107.03374},
  year={2021}
}

@article{huang2024code,
  title={Da-code: Agent data science code generation benchmark for large language models},
  author={Huang, Yiming and Luo, Jianwen and Yu, Yan and Zhang, Yitong and Lei, Fangyu and Wei, Yifan and He, Shizhu and Huang, Lifu and Liu, Xiao and Zhao, Jun and others},
  journal={arXiv preprint arXiv:2410.07331},
  year={2024}
}

@article{egg2025dabstep,
  title={DABstep: Data Agent Benchmark for Multi-step Reasoning},
  author={Egg, Alex and Goyanes, Martin Iglesias and Kingma, Friso and Mora, Andreu and von Werra, Leandro and Wolf, Thomas},
  journal={arXiv preprint arXiv:2506.23719},
  year={2025}
}

@article{lai2025kramabench,
  title={Kramabench: A benchmark for ai systems on data-to-insight pipelines over data lakes},
  author={Lai, Eugenie and Vitagliano, Gerardo and Zhang, Ziyu and Chabra, Om and Sudhir, Sivaprasad and Zeng, Anna and Zabreyko, Anton A and Li, Chenning and Kossmann, Ferdi and Ding, Jialin and others},
  journal={arXiv preprint arXiv:2506.06541},
  year={2025}
}

@inproceedings{lai2023ds,
  title={DS-1000: A natural and reliable benchmark for data science code generation},
  author={Lai, Yuhang and Li, Chengxi and Wang, Yiming and Zhang, Tianyi and Zhong, Ruiqi and Zettlemoyer, Luke and Yih, Wen-tau and Fried, Daniel and Wang, Sida and Yu, Tao},
  booktitle={International Conference on Machine Learning},
  pages={18319--18345},
  year={2023},
  organization={PMLR}
}

@article{huang2023mlagentbench,
  title={Mlagentbench: Evaluating language agents on machine learning experimentation},
  author={Huang, Qian and Vora, Jian and Liang, Percy and Leskovec, Jure},
  journal={arXiv preprint arXiv:2310.03302},
  year={2023}
}

@article{jing2024dsbench,
  title={DSBench: How Far Are Data Science Agents from Becoming Data Science Experts?},
  author={Jing, Liqiang and Huang, Zhehui and Wang, Xiaoyang and Yao, Wenlin and Yu, Wenhao and Ma, Kaixin and Zhang, Hongming and Du, Xinya and Yu, Dong},
  journal={arXiv preprint arXiv:2409.07703},
  year={2024}
}

@inproceedings{zhang2024benchmarking,
  title={Benchmarking data science agents},
  author={Zhang, Yuge and Jiang, Qiyang and XingyuHan, XingyuHan and Chen, Nan and Yang, Yuqing and Ren, Kan},
  booktitle={Proceedings of the 62nd Annual Meeting of the Association for Computational Linguistics (Volume 1: Long Papers)},
  pages={5677--5700},
  year={2024}
}

@inproceedings{yin2023natural,
  title={Natural language to code generation in interactive data science notebooks},
  author={Yin, Pengcheng and Li, Wen-Ding and Xiao, Kefan and Rao, Abhishek and Wen, Yeming and Shi, Kensen and Howland, Joshua and Bailey, Paige and Catasta, Michele and Michalewski, Henryk and others},
  booktitle={Proceedings of the 61st Annual Meeting of the Association for Computational Linguistics (Volume 1: Long Papers)},
  pages={126--173},
  year={2023}
}

@article{rahman2025llm,
  title={Llm-based data science agents: A survey of capabilities, challenges, and future directions},
  author={Rahman, Mizanur and Bhuiyan, Amran and Islam, Mohammed Saidul and Laskar, Md Tahmid Rahman and Mahbub, Ridwan and Masry, Ahmed and Joty, Shafiq and Hoque, Enamul},
  journal={arXiv preprint arXiv:2510.04023},
  year={2025}
}

@article{jiang2024survey,
  title={A survey on large language models for code generation},
  author={Jiang, Juyong and Wang, Fan and Shen, Jiasi and Kim, Sungju and Kim, Sunghun},
  journal={arXiv preprint arXiv:2406.00515},
  year={2024}
}

@article{wang2024gui,
  title={Gui agents with foundation models: A comprehensive survey},
  author={Wang, Shuai and Liu, Weiwen and Chen, Jingxuan and Zhou, Yuqi and Gan, Weinan and Zeng, Xingshan and Che, Yuhan and Yu, Shuai and Hao, Xinlong and Shao, Kun and others},
  journal={arXiv preprint arXiv:2411.04890},
  year={2024}
}

@inproceedings{rahman2025text2vis,
  title={Text2vis: A challenging and diverse benchmark for generating multimodal visualizations from text},
  author={Rahman, Mizanur and Laskar, Md Tahmid Rahman and Joty, Shafiq and Hoque, Enamul},
  booktitle={Proceedings of the 2025 Conference on Empirical Methods in Natural Language Processing},
  pages={31837--31862},
  year={2025}
}

@article{yao2022webshop,
  title={Webshop: Towards scalable real-world web interaction with grounded language agents},
  author={Yao, Shunyu and Chen, Howard and Yang, John and Narasimhan, Karthik},
  journal={Advances in Neural Information Processing Systems},
  volume={35},
  pages={20744--20757},
  year={2022}
}

@article{deng2023mind2web,
  title={Mind2web: Towards a generalist agent for the web},
  author={Deng, Xiang and Gu, Yu and Zheng, Boyuan and Chen, Shijie and Stevens, Sam and Wang, Boshi and Sun, Huan and Su, Yu},
  journal={Advances in Neural Information Processing Systems},
  volume={36},
  pages={28091--28114},
  year={2023}
}

@inproceedings{koh2024visualwebarena,
  title={Visualwebarena: Evaluating multimodal agents on realistic visual web tasks},
  author={Koh, Jing Yu and Lo, Robert and Jang, Lawrence and Duvvur, Vikram and Lim, Ming and Huang, Po-Yu and Neubig, Graham and Zhou, Shuyan and Salakhutdinov, Russ and Fried, Daniel},
  booktitle={Proceedings of the 62nd Annual Meeting of the Association for Computational Linguistics (Volume 1: Long Papers)},
  pages={881--905},
  year={2024}
}

@article{wang2024officebench,
  title={Officebench: Benchmarking language agents across multiple applications for office automation},
  author={Wang, Zilong and Cui, Yuedong and Zhong, Li and Zhang, Zimin and Yin, Da and Lin, Bill Yuchen and Shang, Jingbo},
  journal={arXiv preprint arXiv:2407.19056},
  year={2024}
}

@inproceedings{li2025screenspot,
  title={Screenspot-pro: Gui grounding for professional high-resolution computer use},
  author={Li, Kaixin and Meng, Ziyang and Lin, Hongzhan and Luo, Ziyang and Tian, Yuchen and Ma, Jing and Huang, Zhiyong and Chua, Tat-Seng},
  booktitle={Proceedings of the 33rd ACM International Conference on Multimedia},
  pages={8778--8786},
  year={2025}
}

@article{zhou2023webarena,
  title={Webarena: A realistic web environment for building autonomous agents},
  author={Zhou, Shuyan and Xu, Frank F and Zhu, Hao and Zhou, Xuhui and Lo, Robert and Sridhar, Abishek and Cheng, Xianyi and Ou, Tianyue and Bisk, Yonatan and Fried, Daniel and others},
  journal={arXiv preprint arXiv:2307.13854},
  year={2023}
}

@article{rawles2024androidworld,
  title={Androidworld: A dynamic benchmarking environment for autonomous agents},
  author={Rawles, Christopher and Clinckemaillie, Sarah and Chang, Yifan and Waltz, Jonathan and Lau, Gabrielle and Fair, Marybeth and Li, Alice and Bishop, William and Li, Wei and Campbell-Ajala, Folawiyo and others},
  journal={arXiv preprint arXiv:2405.14573},
  year={2024}
}

@article{sarker2021data,
  title={Data science and analytics: an overview from data-driven smart computing, decision-making and applications perspective},
  author={Sarker, Iqbal H},
  journal={SN Computer Science},
  volume={2},
  number={5},
  pages={377},
  year={2021},
  publisher={Springer}
}

@article{adeniran2024role,
  title={The role of data science in transforming business operations: Case studies from enterprises},
  author={Adeniran, Ibrahim Adedeji and Efunniyi, Christianah Pelumi and Osundare, Olajide Soji and Abhulimen, Angela Omozele and OneAdvanced, UK},
  journal={Computer Science \& IT Research Journal},
  volume={5},
  number={8},
  pages={2026--2039},
  year={2024}
}

@article{cao2017data,
  title={Data science: a comprehensive overview},
  author={Cao, Longbing},
  journal={ACM Computing Surveys (CSUR)},
  volume={50},
  number={3},
  pages={1--42},
  year={2017},
  publisher={ACM New York, NY, USA}
}

@book{dale2022data,
  title={Data Visualization with Python and JavaScript: Scrape, Clean, Explore, and Transform Your Data},
  author={Dale, Kyran},
  year={2022},
  publisher={" O'Reilly Media, Inc."}
}

@article{lei2024spider,
  title={Spider 2.0: Evaluating language models on real-world enterprise text-to-sql workflows},
  author={Lei, Fangyu and Chen, Jixuan and Ye, Yuxiao and Cao, Ruisheng and Shin, Dongchan and Su, Hongjin and Suo, Zhaoqing and Gao, Hongcheng and Hu, Wenjing and Yin, Pengcheng and others},
  journal={arXiv preprint arXiv:2411.07763},
  year={2024}
}

@article{chen2024viseval,
  title={Viseval: A benchmark for data visualization in the era of large language models},
  author={Chen, Nan and Zhang, Yuge and Xu, Jiahang and Ren, Kan and Yang, Yuqing},
  journal={IEEE Transactions on Visualization and Computer Graphics},
  year={2024},
  publisher={IEEE}
}

@article{liu2023your,
  title={Is your code generated by chatgpt really correct? rigorous evaluation of large language models for code generation},
  author={Liu, Jiawei and Xia, Chunqiu Steven and Wang, Yuyao and Zhang, Lingming},
  journal={Advances in Neural Information Processing Systems},
  volume={36},
  pages={21558--21572},
  year={2023}
}

@inproceedings{hong2025data,
  title={Data interpreter: An {LLM} agent for data science},
  author={Hong, Sirui and Lin, Yizhang and Liu, Bang and Liu, Bangbang and Wu, Binhao and Zhang, Ceyao and Li, Danyang and Chen, Jiaqi and Zhang, Jiayi and Wang, Jinlin and others},
  booktitle={Findings of the Association for Computational Linguistics: ACL 2025},
  pages={19796--19821},
  year={2025}
}

@article{tang2025survey,
  title={A survey on (m) llm-based gui agents},
  author={Tang, Fei and Xu, Haolei and Zhang, Hang and Chen, Siqi and Wu, Xingyu and Shen, Yongliang and Zhang, Wenqi and Hou, Guiyang and Tan, Zeqi and Yan, Yuchen and others},
  journal={arXiv preprint arXiv:2504.13865},
  year={2025}
}

@book{wickham2023r,
  title={R for data science: import, tidy, transform, visualize, and model data},
  author={Wickham, Hadley and {\c{C}}etinkaya-Rundel, Mine and Grolemund, Garrett},
  year={2023},
  publisher={" O'Reilly Media, Inc."}
}

@article{donoho201750,
  title={50 years of data science},
  author={Donoho, David},
  journal={Journal of Computational and Graphical Statistics},
  volume={26},
  number={4},
  pages={745--766},
  year={2017},
  publisher={Taylor \& Francis}
}

@article{kartha2025dashboardqa,
  title={DashboardQA: Benchmarking Multimodal Agents for Question Answering on Interactive Dashboards},
  author={Kartha, Aaryaman and Masry, Ahmed and Islam, Mohammed Saidul and Lang, Thinh and Rahman, Shadikur and Mahbub, Ridwan and Rahman, Mizanur and Ahmed, Mahir and Parvez, Md Rizwan and Hoque, Enamul and others},
  journal={arXiv preprint arXiv:2508.17398},
  year={2025}
}

@article{feizi2025grounding,
  title={Grounding Computer Use Agents on Human Demonstrations},
  author={Feizi, Aarash and Nayak, Shravan and Jian, Xiangru and Lin, Kevin Qinghong and Li, Kaixin and Awal, Rabiul and L{\`u}, Xing Han and Obando-Ceron, Johan and Rodriguez, Juan A and Chapados, Nicolas and others},
  journal={arXiv preprint arXiv:2511.07332},
  year={2025}
}

@inproceedings{hong2024cogagent,
  title={Cogagent: A visual language model for gui agents},
  author={Hong, Wenyi and Wang, Weihan and Lv, Qingsong and Xu, Jiazheng and Yu, Wenmeng and Ji, Junhui and Wang, Yan and Wang, Zihan and Dong, Yuxiao and Ding, Ming and others},
  booktitle={Proceedings of the IEEE/CVF Conference on Computer Vision and Pattern Recognition},
  pages={14281--14290},
  year={2024}
}

@inproceedings{lin2025showui,
  title={Showui: One vision-language-action model for gui visual agent},
  author={Lin, Kevin Qinghong and Li, Linjie and Gao, Difei and Yang, Zhengyuan and Wu, Shiwei and Bai, Zechen and Lei, Stan Weixian and Wang, Lijuan and Shou, Mike Zheng},
  booktitle={Proceedings of the Computer Vision and Pattern Recognition Conference},
  pages={19498--19508},
  year={2025}
}

@inproceedings{you2024ferret,
  title={Ferret-ui: Grounded mobile ui understanding with multimodal llms},
  author={You, Keen and Zhang, Haotian and Schoop, Eldon and Weers, Floris and Swearngin, Amanda and Nichols, Jeffrey and Yang, Yinfei and Gan, Zhe},
  booktitle={European Conference on Computer Vision},
  pages={240--255},
  year={2024},
  organization={Springer}
}

@article{wu2024atlas,
  title={Os-atlas: A foundation action model for generalist gui agents},
  author={Wu, Zhiyong and Wu, Zhenyu and Xu, Fangzhi and Wang, Yian and Sun, Qiushi and Jia, Chengyou and Cheng, Kanzhi and Ding, Zichen and Chen, Liheng and Liang, Paul Pu and others},
  journal={arXiv preprint arXiv:2410.23218},
  year={2024}
}

@article{xie2025scaling,
  title={Scaling Computer-Use Grounding via User Interface Decomposition and Synthesis},
  author={Xie, Tianbao and Deng, Jiaqi and Li, Xiaochuan and Yang, Junlin and Wu, Haoyuan and Chen, Jixuan and Hu, Wenjing and Wang, Xinyuan and Xu, Yuhui and Wang, Zekun and others},
  journal={arXiv preprint arXiv:2505.13227},
  year={2025}
}

@article{luo2025gui,
  title={Gui-r1: A generalist r1-style vision-language action model for gui agents},
  author={Luo, Run and Wang, Lu and He, Wanwei and Chen, Longze and Li, Jiaming and Xia, Xiaobo},
  journal={arXiv preprint arXiv:2504.10458},
  year={2025}
}

@article{tang2025gui,
  title={GUI-G2: Gaussian Reward Modeling for GUI Grounding},
  author={Tang, Fei and Gu, Zhangxuan and Lu, Zhengxi and Liu, Xuyang and Shen, Shuheng and Meng, Changhua and Wang, Wen and Zhang, Wenqi and Shen, Yongliang and Lu, Weiming and others},
  journal={arXiv preprint arXiv:2507.15846},
  year={2025}
}

@article{liu2025infigui,
  title={Infigui-g1: Advancing gui grounding with adaptive exploration policy optimization},
  author={Liu, Yuhang and Liu, Zeyu and Zhu, Shuanghe and Li, Pengxiang and Xie, Congkai and Wang, Jiasheng and Hu, Xueyu and Han, Xiaotian and Yuan, Jianbo and Wang, Xinyao and others},
  journal={arXiv preprint arXiv:2508.05731},
  year={2025}
}

@misc{openai2025computer,
  title={Computer-Using Agent},
  author={OpenAI},
  year={2024},
  url={https://cdn.openai.com/pdf/7ef17d82-96bf-4dd1-9df2-228f7f377a29/the-state-of-enterprise-ai_2025-report.pdf},
  note={Accessed: 2025-12-14}
}

@techreport{anaconda2025state,
  title={2025 State of Data Science: Moving From Hype Toward Maturity},
  author={{Anaconda}},
  year={2025},
  institution={Anaconda, Inc.},
  url={https://www.anaconda.com/wp-content/uploads/2025/03/Anaconda_SODS.pdf},
  note={Accessed: 2025-12-16}
}

@article{wang2025opencua,
  title={Opencua: Open foundations for computer-use agents},
  author={Wang, Xinyuan and Wang, Bowen and Lu, Dunjie and Yang, Junlin and Xie, Tianbao and Wang, Junli and Deng, Jiaqi and Guo, Xiaole and Xu, Yiheng and Wu, Chen Henry and others},
  journal={arXiv preprint arXiv:2508.09123},
  year={2025}
}
